\documentclass[conference]{IEEEtran}
\usepackage{times}

\usepackage[numbers]{natbib}
\usepackage[pagebackref=true,breaklinks=true,bookmarks=true,colorlinks]{hyperref}
\usepackage{url}
\usepackage{tikz} 
\usepackage{amsmath}
\usepackage{amssymb}
\usepackage{multirow}
\usepackage{multicol}
\usepackage{mathrsfs}
\usepackage{etoolbox}
\usepackage{xspace}
\usepackage{xcolor}
\usepackage{graphicx}
\usepackage{caption}
\usepackage{subcaption}
\usepackage{adjustbox}
\usepackage{xfrac}
\usepackage{listings}
\usepackage{float}
\usepackage{booktabs}
\usepackage{bm}
\usepackage{amsfonts}

\definecolor{deepblue}{HTML}{047bff}

\newcommand{\etc}{\textit{etc}}
\newcommand{\etal}{\textit{et al}. }

\newcommand{\eg}{\textit{e}.\textit{g}., }
\newcommand{\teleop}{TactAR\xspace}
\newcommand{\alg}{Reactive Diffusion Policy\xspace}
\newcommand{\ourwebsite}[0]{\href{https://reactive-diffusion-policy.github.io/}{reactive-diffusion-policy.github.io}}

\begin{document}

\title{Reactive Diffusion Policy: \\ \LARGE{Slow-Fast Visual-Tactile Policy Learning for Contact-Rich Manipulation}}
\vspace{-0.2in}

\author{Han Xue$^{1*}$\quad Jieji Ren$^{1*}$\quad Wendi Chen$^{1*}$\\ Gu Zhang$^{234\dagger}$\quad Yuan Fang$^{1\dagger}$\quad Guoying Gu$^{1}$\quad Huazhe Xu$^{234\ddagger}$\quad Cewu Lu$^{15\ddagger}$\vspace{0.03in}\\

$^1$Shanghai Jiao Tong University\quad$^2$Tsinghua University, IIIS\quad$^3$Shanghai Qi Zhi Institute\quad \\$^4$Shanghai AI Laboratory
$^5$Shanghai Innovation Institute \vspace{0.03in}\\\quad$^*$Equal contribution\quad$^\dagger$Equal contribution\quad  $^\ddagger$Equal advising\vspace{0.1in}\\

\href{https://reactive-diffusion-policy.github.io/}{\color{deepblue}\textbf{reactive-diffusion-policy.github.io}\xspace}\vspace{-0.1in}}


\IEEEpeerreviewmaketitle
\twocolumn[{%
    \renewcommand\twocolumn[1][]{#1}%
        \maketitle
        \vspace{-5mm}
	\begin{center}

\vspace{-0cm}
    \includegraphics[width=18cm]{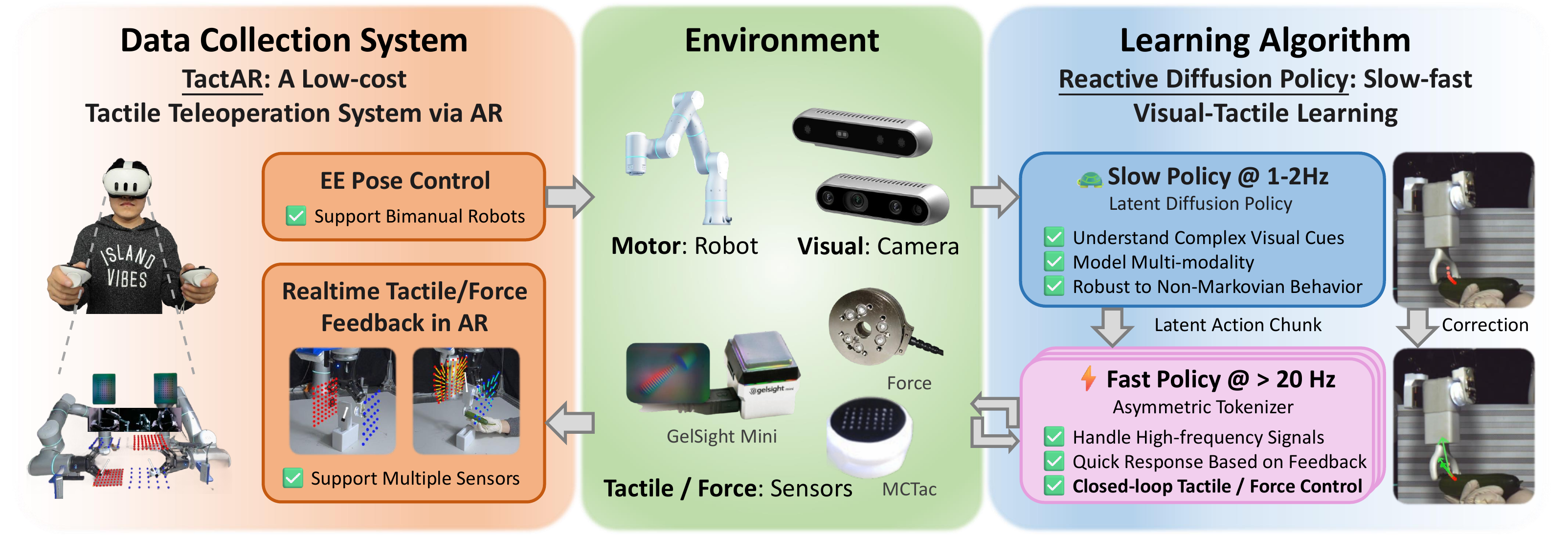}
    \captionof{figure}{\textbf{\teleop} is a low-cost and versatile teleoperation system which can provide real-time tactile / force feedback via Augmented Reality (AR). \textbf{\alg (RDP)} is a slow-fast imitation learning algorithm that can model complex action trajectories with a slow policy network and achieve closed-loop control based on high-frequency tactile / force feedback with a fast policy network. }
    \label{fig:teaser}
    \end{center}
}]

\begin{abstract}
Humans can accomplish complex contact-rich tasks using vision and touch, with highly reactive capabilities such as fast response to external changes and adaptive control of contact forces; however, this remains challenging for robots. Existing visual imitation learning (IL) approaches rely on action chunking to model complex behaviors, which lacks the ability to respond instantly to real-time tactile feedback during the chunk execution. Furthermore, most teleoperation systems struggle to provide fine-grained tactile / force feedback, which limits the range of tasks that can be performed. To address these challenges, we introduce TactAR, a low-cost teleoperation system that provides real-time tactile feedback through Augmented Reality (AR), along with Reactive Diffusion Policy (RDP), a novel slow-fast visual-tactile imitation learning algorithm for learning contact-rich manipulation skills. RDP employs a two-level hierarchy: (1) a slow latent diffusion policy for predicting high-level action chunks in latent space at low frequency, (2) a fast asymmetric tokenizer for closed-loop tactile feedback control at high frequency. This design enables both complex trajectory modeling and quick reactive behavior within a unified framework. Through extensive evaluation across three challenging contact-rich tasks, RDP significantly improves performance  compared to state-of-the-art visual IL baselines. Furthermore, experiments show that RDP is applicable across different tactile / force sensors. Code and videos are available on \ourwebsite. 
\end{abstract}

\section{Introduction}
Humans are capable of performing numerous precise contact-rich tasks (\eg peeling vegetables) in daily life by employing both vision and touch. However, these contact-rich tasks that appear simple to humans can be quite challenging for robots. Some research work in neuroscience \cite{johansson2009coding, flanagan2006control, johansson2007sensory} has indicated that when humans engage in contact-rich tasks, the process can be divided into two components: \textbf{1) feedforward / predictive action} and \textbf{2) closed-loop fine-tuning} based on sensory feedback, such as tactile signals. Inspired by this, we aim to develop a robot learning system capable of emulating human control patterns when performing intricate contact-rich tasks.

In recent years, visual imitation learning (IL) methods \cite{diffusion_policy, act} have demonstrated strong performance in various real-world tasks.  Representative works in this area, such as Diffusion Policy \cite{diffusion_policy}, ACT \cite{act} and $ \pi_0 $\cite{black2024pi_0}, have employed \textbf{action chunking} to mitigate compounding errors in long sequences and improve temporal consistency. Action chunking also effectively models non-Markovian behaviors commonly found in human demonstrations, such as pauses or oscillatory motion. However, these methods operate in an \textbf{open-loop} state during the execution of action chunks, which makes the policy unable to respond instantly to environment changes in contact-rich tasks. Furthermore, the absence of tactile input significantly limits the capabilities of these methods. As a result, these IL methods are limited to low-precision tasks (\eg pick-and-place, push-pull), which do not require precise force control or fast response. 

In order to compensate for the limitations of purely visual input, numerous approaches \cite{hato, huang20243dvitac, pattabiraman2024learning, yu2024mimictouch, jones2025beyond} have explored the integration of tactile input into imitation learning policies. However, most of these research works focus only on the observation level, using tactile input to provide additional information such as visual occlusion or the determination of contact state. At the action level, these methods still rely on conventional action chunking for action prediction, limiting their ability to respond quickly during the execution of the chunk. Furthermore, most approaches utilize traditional teleoperation to collect human demonstration data, making it challenging to obtain high-quality action data with fine-grained tactile feedback. This limitation also limits the types of tasks that these methods can effectively perform. 

In this work, we propose two critical components to solve the above issues of visual-tactile imitation learning: 
\begin{itemize}
    \item  A \textbf{teleoperation system} called \textbf{TactAR} which can provide fine-grained tactile / force feedback in real time through Augmented Reality (AR).
    \item An \textbf{imitation learning algorithm} called \textbf{Reactive Diffusion Policy (RDP)} that retains the advantages of action chunking while enabling high-frequency closed-loop control based on tactile signals during the execution of each chunk. 
\end{itemize}

In our proposed \teleop teleoperation system, we use Meta Quest3 to provide real-time tactile / force feedback via Augmented Reality (AR). We use 3D deformation field as the unified representation for tactile / force feedback (Fig. \ref{fig:teaser} left) of different sensors. We render the 3D deformation field in AR and attach it to the robot end-effector in virtual space. Our system also supports camera streaming of tactile sensors and RGB cameras in AR. In this way, the user can get rich contact information during teleoperation including the tactile image, normal force, shear force, and torsional torques. Our \teleop system is designed with an emphasis on versatility and accessibility: (1) it supports multiple types of tactile / force sensors; (2) it can be easily deployed in different robot embodiments; (3) it is very cost-effective (\$500 for Meta Quest3). 

To leverage the high-quality visual tactile data collected by the \teleop system, we propose an imitation learning algorithm called \alg(RDP) (Fig. \ref{fig:teaser} right).  Inspired by human control strategies in contact-rich tasks, we adopt a ``slow-fast" hierarchical control pipeline. We use a slow network \textbf{(latent diffusion policy, LDP)} to act like a \textbf{neural planner} and predict a high-level action chunk in latent space at low frequency (1-2 Hz), which is analogous to the \textbf{predictive action}. Then we use a fast network \textbf{(Asymmetric Tokenizer, AT)} to act like a \textbf{learnable impedance controller} and finetune the latent action chunk based on high-frequency tactile feedback (20-30 Hz), which is analogous to the \textbf{closed-loop finetuning}. This hierarchical structure allows the slow network to preserve its ability to model complex and non-Markovian actions via the diffusion model \cite{diffusion_policy} and action chunking, while the fast network achieves closed-loop control with real-time tactile feedback for precise force control and quick response.

We have conducted experiments on three challenging contact-rich tasks, which can evaluate the model performance on the following aspects: (1) precision, (2) precise and adaptive force control, (3) fast response under disturbances, and (4) bimanual coordination. Real-world experiments have shown that our \alg algorithm can model complex actions while maintaining very fast reactive behavior, achieving a significant performance improvement ($>35\%$) in three tasks compared to state-of-the-art IL baselines. We have also conducted user studies and quantitative analyses to validate the high quality of the data collected by \teleop teleoperation system. Furthermore, we conduct cross-sensor experiments demonstrating that RDP is applicable across various tactile / force sensors (GelSight Mini\cite{gelsight_mini}, MCTac \cite{mctac} and built-in joint torque sensors \cite{flexiv_rizon4}). The code of the \teleop teleoperation system and \alg algorithm are available on \ourwebsite. 

\begin{figure*}[!ht]
    \centering
    \includegraphics[width=\linewidth]{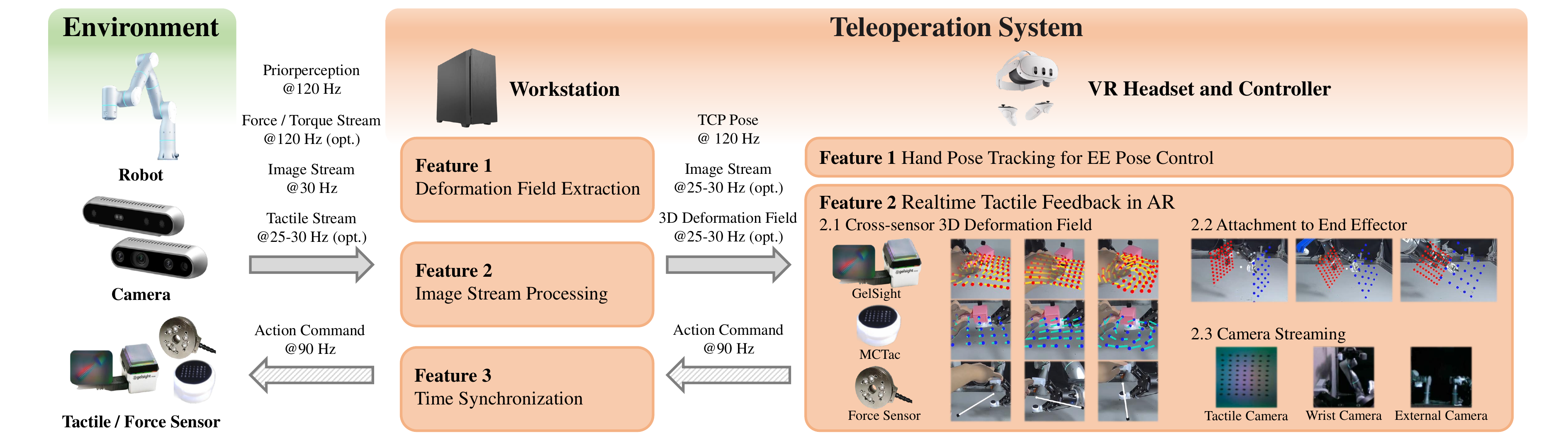}
    \caption{Overview of \textbf{\teleop}\xspace teleoperation system. It can provide real-time tactile / force feedback via Augmented Reality (AR). The tactile feedback is represented as the 3D deformation field, which is a universal representation applicable to multiple different tactile / force sensors. The 3D deformation field is rendered and ``attached" to the robot end-effector in AR, which makes the user perceive the rich contact information in 3D space. \teleop also support real-time streaming for multiple RGB cameras and optical tactile sensors. Please see the video in the supplementary file for more details.}
    \label{fig:teleop_system}
    \vspace{-5mm}
\end{figure*}

\section{Related Work}

\subsection{Tactile / Force Sensing Hardware in Robotics}
Many robotics tasks require physical interaction with the world. Tactile sensors and force sensors can provide richer information about the contact physics compared to RGB cameras. Each type of sensors have their advantages and limitations. Force sensors mounted on the robot end effector or joint torque sensors on the robot arm can directly obtain force / torque readings and are less prone to damage. Nevertheless, force and torque sensors are prone to signal noise, particularly during rapid movement, and they also tend to be relatively expensive.

The tactile sensors can be classified as electrical sensors and optical sensors. Electrical sensors use capacitative \cite{glauser2019deformation,wu2020capacitivo,xu2015stretch,xu2024cushsense}, resistive \cite{sundaram2019learning,bhattacharjee2013tactile,stassi2014flexible}, MEMS \cite{wettels2008biomimetic} or magnetic particles \cite{bhirangi2021reskin,hellebrekers2019soft,tomo2018new,yan2021soft,bhirangi2024anyskin} to sense contact. Such sensors are usually more compact and thin, but have lower spatial resolution and are complex to manufacture. Only some electrical tactile sensors can directly output both normal force and shear force \cite{bhirangi2024anyskin}, but rely on a complex calibration process with force/torque sensors. 

Optical tactile sensors \cite{yuan2017gelsight,taylor2022gelslim,lambeta2020digit,mctac,lin20239dtact, lin2023dtact, kuppuswamy2020soft} such as GelSight~\cite{yuan2017gelsight}, GelSlim~\cite{taylor2022gelslim}, MCTac\cite{mctac} and 9DTact~\cite{lin20239dtact, lin2023dtact} are another stream of works; they capture high-resolution images with cameras to track the deformation of gels; optical sensors can also be equipped with markers dots to better track the normal and shear deformation field of the gel surface. Compared to electrical sensors or force / torque sensors, optical tactile sensors are easier to fabricate, less expensive and have lower signal noise. Force and torque information can be implicitly represented on the shear field gel surface but they rely on complex calibration process \cite{zhang2019effective} to indirectly calculate force/torque values. In this paper, we use two different optical tactile sensors (GelSight Mini \cite{gelsight_mini} and MCTac\cite{mctac}) and the joint torque sensors of the robot arm to further unleash the potential of tactile / force sensing.

\subsection{Robot Data Collection System}
Teleoperation is a common way to collect expert data with robots. Current teleoperation systems are mainly built upon VR controller \cite{iyer2024open}, hand tracking \cite{qin2023anyteleop} or direct joint mapping  \cite{wu2024gello,fang2024airexo,act} with hardwares.  Most teleoperation systems are soley based on visual feedback, which makes them hard to perform precise contact-rich tasks. 

One stream of haptic teleoperation systems relies on isomorphic hardwares.  Bi-ACT \cite{buamanee2024bi}  and \citet{kobayashi2024alpha} use bilateral control with ALOHA \cite{act} to get force feedback. These teleoperation systems are difficult to deploy on different hardware platforms due to isomorphic hardware designs. Another stream of teleoperation systems use force/torque sensors for haptic feedback. FoAR \cite{he2024foar} uses a haptic teleoperation device \cite{sigma7, fang2024rh20t} to obtain force feedback, but it also has a relatively high cost. ForceMimic \cite{liu2024forcemimic} adds a force sensor on a hand-held device \cite{chi2024universal} to get force feedback, but suffers from the inaccuracy of pose estimation, and thus cannot directly train an end2end policy. ACP \cite{hou2024adaptive} uses a low-stiffness compliance kinesthetic teaching system to provide haptic feedback, but requires complex custom adapters for the end effectors.

Some research works use vibration in VR headset \cite{comp_act, aburub2024learning, zhou2024contact,ding2024bunny} to get tactile feedback, but these systems can only provide very coarse information about contact. Our method combines the advantages of low-cost VR controller and tactile sensing, getting tactile feedback via Augmented Reality, while preserving the accuracy needed for precise contact-rich tasks. In addition, our teleoperation systems are flexible and easy to deploy on different tactile sensors and robot platforms.

\subsection{Visual-tactile Manipulation}
Tactile sensing is crucial in robotic manipulation, compensating for visual occlusions and providing force/torque feedback in complex tasks. While  manipulation with tactile sensing has a long history 
\cite{yu2023precise,robot_synesthesia,guzey2024see,ai2024robopack}, many approaches rely on task-specific modeling \cite{oller2024tactile,ai2024robopack}, hand-designed primitives \cite{yu2023precise,ye2024morpheus}, or hand-crafted rewards \cite{robot_synesthesia}, limiting their generalizability on different tasks. 

Recent advances in visual Imitation Learning (e.g., \cite{diffusion_policy, act,Ze2024DP3}) have shown promise in learning complex tasks in an end-to-end way. 
Several studies \cite{hato, huang20243dvitac,bogert2024built,yu2024mimictouch} have attempted to integrate tactile sensing with imitation learning, but most focus solely on normal force readings, limiting their applicability to versatile tasks.  Only a few works leverage shear force/torque data for imitation learning. MimicTouch \cite{yu2024mimictouch} uses a GelSight \cite{yuan2017gelsight}-based handheld device for offline policy training but omits visual input due to the embodiment gap between human hands and robot grippers, limiting its task versatility. Similarly, Bogert \etal  \cite{bogert2024built} uses GelSlim \cite{sipos2024gelslim}’s shear force field for policy transfer across robot embodiments but also excludes visual input, restricting their approach to simple tasks.  

In contrast, our method combines normal force, shear force, and visual RGB inputs into a unified visual-tactile policy, enabling deployment across a broader range of tasks. By integrating both tactile and visual modalities, our approach overcomes the limitations of prior works and achieves greater versatility in robotic manipulation.

\section{Teleoperation System: \teleop} \label{sec:teleop}
\teleop is an AR-based teleoperation system (see Fig. \ref{fig:teleop_system}) that can provide real-time tactile / force feedback for complex contact-rich tasks. The main features of \teleop's system design are:
\begin{itemize}

\item \textbf{Real-time tactile / force feedback via AR.} The 3D deformation / force field of the tactile / force sensors can be rendered in real time with Augmented Reality (AR). After a simple calibration step, the 3D deformation field will  ``attach" to the robot end-effector in AR, which can provide real-time tactile / force / torque feedbacks to the user.
\item \textbf{Cross-sensor and cross-embodiment deployment.} The system can be deployed cross different types of tactile / force sensors  and different robot arms (single-arm or bimanual).
\item \textbf{Low-cost.} The system do not need specialized or isomorphic hardwares for haptic feedback. It only needs a consumer-level VR headset (Meta Quest3) for teleoperation, which costs only $\$500$.

\end{itemize}

Next, we will introduce critical components and features of \teleop.  

\subsection{3D Deformation Field Extraction}\label{sec:marker_field_extraction}
\begin{figure}[h]
    \centering
    \includegraphics[width=\linewidth]{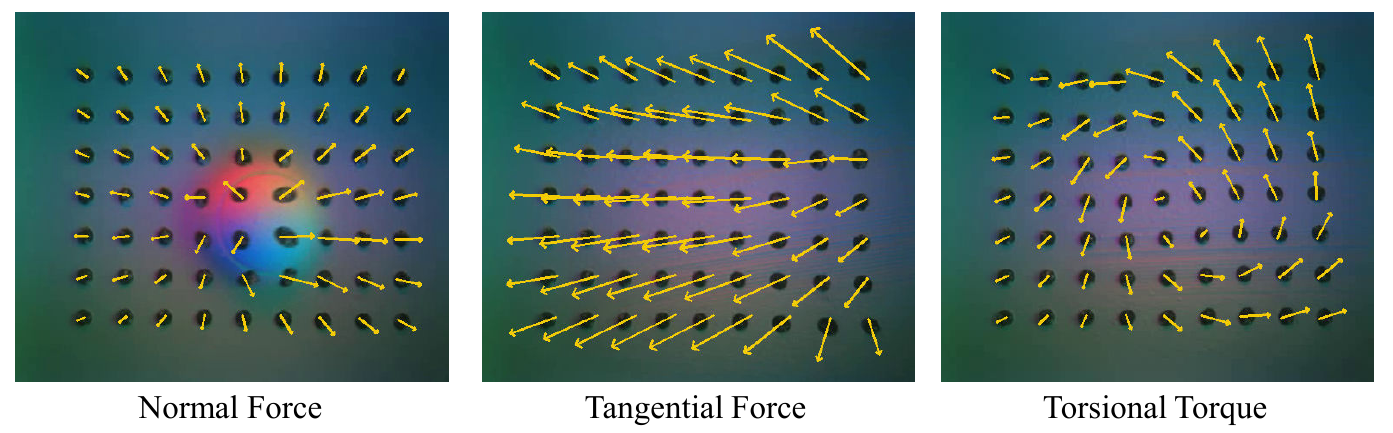}
    \caption{Examples of marker deformation field in GelSight Mini \cite{gelsight_mini} during different contact modes.}
    \label{fig:deformation_field}
    \vspace{-3mm}
\end{figure}

The gel surface's marker array (Fig. \ref{fig:deformation_field}) captures rich contact information, but deriving forces from 2D optical flow requires complex calibration with expensive sensors. To improve accessibility, we instead visualize the 3D deformation field. From tactile images $I_t$, we extract normalized marker positions $D_t$ using OpenCV \cite{opencv}. We use a score-based tracking algorithm \cite{gelsight_sdk} to calculate 2D optical flow between the initial frame $D_0$ and the current frame $D_t$:

\begin{equation}
    F_t = [\mathbf{d}_x, \mathbf{d}_y]=Flow(D_{0}, D_{t})
    \label{eq:flow_2d}
\end{equation}

The 3D deformation field $V_t=[\mathbf{d}_x, \mathbf{d}_y, \mathbf{o}_z]$ (with z-offset $\mathbf{o_z}$) is then rendered in AR. For force sensors, we directly visualize $V_t=[\mathbf{f}_x, \mathbf{f}_y, \mathbf{f}_z]$.
\subsection{Real-time Tactile / Force Feedback Rendering in AR}
\begin{figure}[h]
    \centering
    \includegraphics[width=\linewidth]{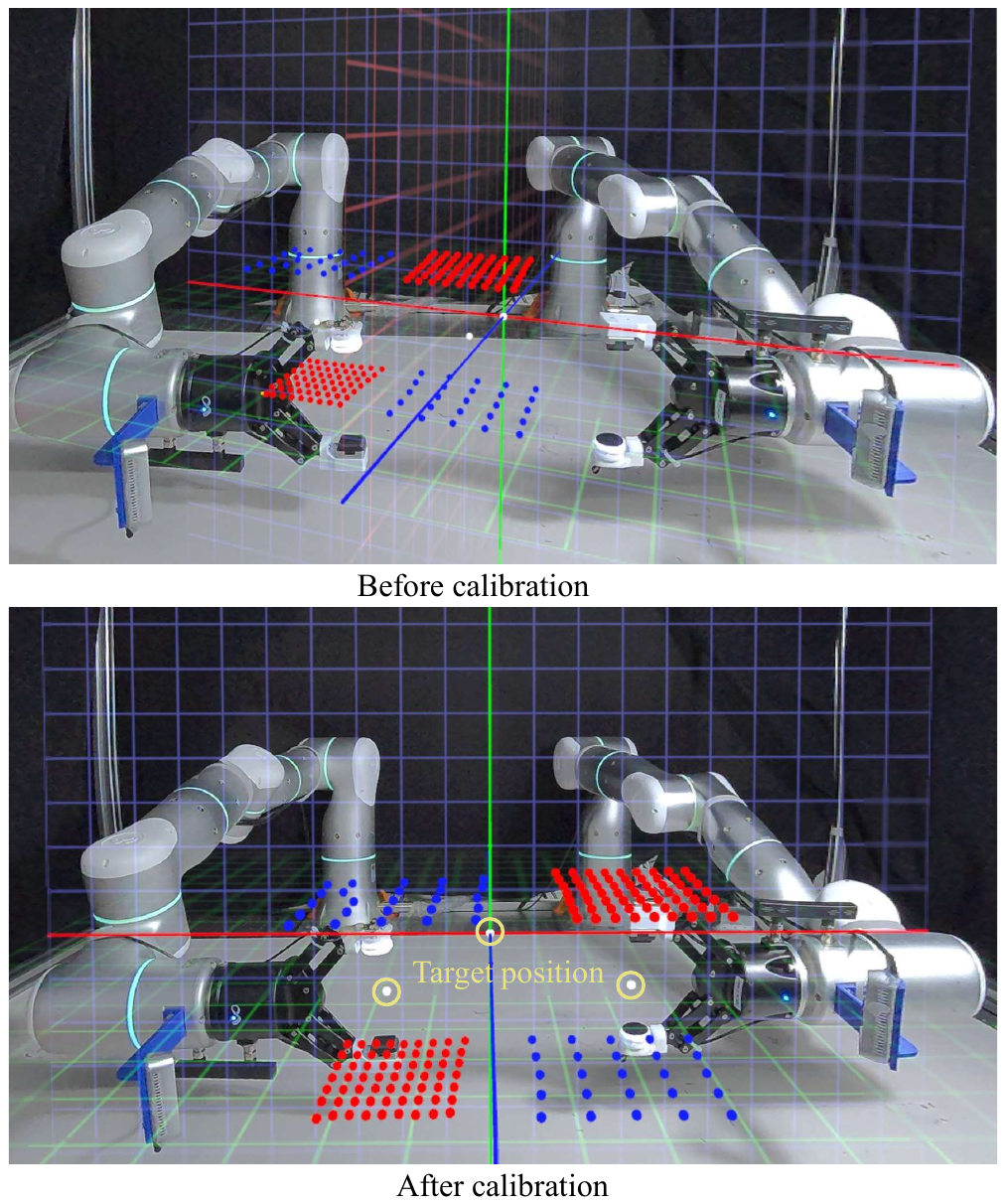}
    \caption{Calibration process in AR. The user adjust the translation and rotation of the virtual coordinate system such that it can align with the pre-defined TCP position (the white sphere) and the origin of the world coordinate system.}
    \label{fig:calibration}
    \vspace{-3mm}
\end{figure}
Our \teleop system employs Meta Quest3's  color pass-through mode through Unity to create an AR environment. The native SLAM algorithm in Quest3 tracks headset/controller poses, with initial coordinate alignment between AR and robot spaces via a simple calibration process (Fig. \ref{fig:calibration}). The ROS2-based architecture (Fig. \ref{fig:teleop_system}) synchronizes data streams from tactile/force sensors, robot controllers, and RGB cameras. Tactile/force information (3D deformation field $V_t$) is transformed using real-time robot TCP poses and rendered in AR space.  Optionally, our \teleop system also supports real-time streaming of multi-view RGB cameras (see in Fig. \ref{fig:teleop_system}) and tactile cameras for more immersive teleoperation experience.  Please see Appendix \ref{sec:supp_tactar_details} and our website for the system latency and more details of \teleop.

\subsection{Versatility and Accessibility}
Our \teleop system is designed to be versatile and easily accessible with the following properties: 
\begin{itemize}
    \item \textbf{Cross-sensor.} Our \teleop system uses the 3D deformation field as the unified representation for tactile / force feedbacks, which can be applied on many different tactile / force sensors. In this paper, we the following sensors for experiments: (1) GelSight Mini \cite{gelsight_mini} (Robotics Package) optical tactile sensor. (2) Our improved MCTac \cite{mctac} optical tactile sensor. (3) Built-in joint torque sensors in Flexiv Rizon \cite{flexiv_rizon4} robot arm.
    Our \teleop system is highly modularized and can even support AR visualization with different tactile / force sensors at the same time (see Fig. \ref{fig:teaser}). Theoretically, our teleoperation system is also compatible with electrically tactile sensors like AnySkin \cite{bhirangi2024anyskin}, whose data types can typically be represented as a sparse 3D force field.
    
    \item \textbf{Cross-embodiment deployment}. Our \teleop system can easily be deployed in different robot arms and grippers. The tactile feedbacks in AR only need the robot TCP pose and the 3D deformation field, and the robot arm uses TCP control. We also support both single-arm and bimanual arm control. Thus, \teleop will not be limited by specific hardware configuration parameters and degrees of freedom. Compared to other haptic teleoperation systems based on isomorphic hardware\cite{comp_act, buamanee2024bi}, our system only needs one Meta Quest3 VR headset, which greatly reduces the reproducibility difficulty.
    \item \textbf{Low cost}. Our \teleop system is built with low-cost hardwares. The Meta Quest3 VR heaset used for teleoperation and AR feedbacks costs $\$499$.  Our experiments use two different optical tactile sensors: (1) GelSight Mini Robotics Package \cite{gelsight_mini} which is a commercialized product costs $\$549$. (2) Our customized MCTac \cite{mctac} costs around $\$50$ in lab fabrication, and there is large potential for lower cost in industrial manufacturing in the future. Compared to force/torque sensors like ATI mini45 \cite{ati_mini45} which costs about $\$3000$, the optical tactile sensors offer a significant cost advantage. 
\end{itemize}

\begin{figure}[!ht]
    \centering    
    \includegraphics[width=\columnwidth]{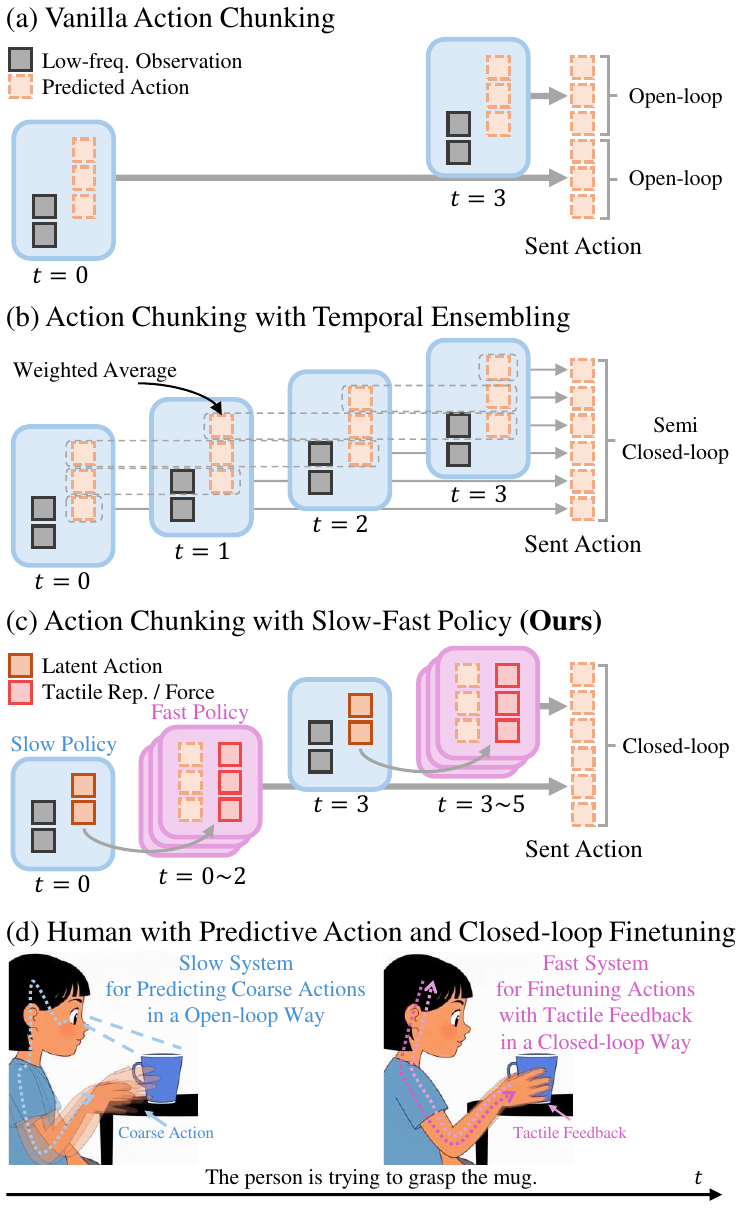}
    \caption{Comparison among various pipelines. (a) Vanilla action chunking \cite{diffusion_policy} with open-loop control during the chunk execution. (b) Action chunking enhanced with temporal ensembling \cite{act,hato} for semi-closed-loop control. (c) Our slow-fast inference pipeline, showcasing closed-loop capabilities with fast responsive adjustments. (d) Human control patterns in contact-rich tasks.}
    \label{fig:pipeline_comparison}
    \vspace{-7mm}
\end{figure}

\section{Learning Algorithm: \alg} \label{sec:algorithm}
In this section, we will introduce \alg ~(RDP), which is a slow-fast imitation learning algorithm that can respond instantly to tactile / force feedback with a fast network while simultaneously maintaining the powerful modeling capabilities of diffusion with a slow network.  

\subsection{Tactile / Force Representation}
\label{sec:tactile_representation}

For optical tactile sensors (e.g. Gelsight Mini \cite{gelsight_mini}), we use low-dimensional representation  generated by Principal Component Analysis~(PCA) on the marker deformation field $\mathbf{F}$ in Eq. \ref{eq:flow_2d}. The use of PCA feature makes the model more robust to tracking errors and noise of marker deformation field. In addition, PCA feature are more robust to texture and lighting changes due to damage or gel replacements. Please see Appendix \ref{sec:supp_tactile_pca} and Appendix \ref{sec:supp_dataset} for more details of the tactile representation and the tactile dataset for PCA.

For force representation, we simply concatenate the 6-D wrench (force + torque) into the observation vector.

\subsection{Slow-Fast Policy Learning}

\begin{figure*}[!ht]
    \centering
    \includegraphics[width=\textwidth]{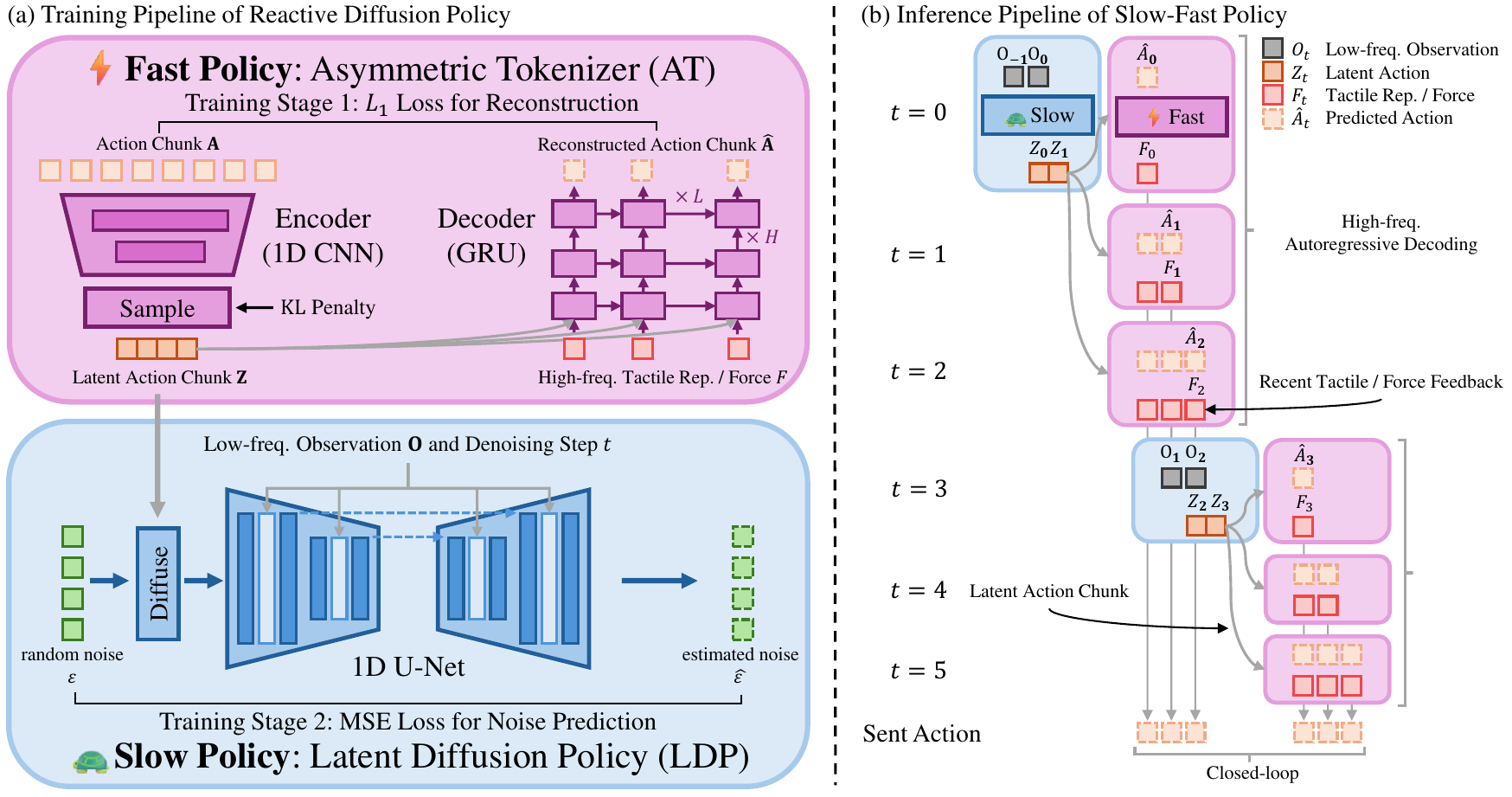}
    \caption{Overview of \textbf{\alg(RDP)} framework. (a) The training pipeline of RDP, comprising the first stage for training the fast policy (Asymmetric Tokenizer) and the second stage for training the slow policy (Latent Diffusion Policy). (b) The inference pipeline of RDP. The slow policy leverages low-frequency observations for modeling complex behaviors with diffusion and action chunking. The fast policy enables closed-loop control by using high-frequency tactile / force input and fine-tuning the latent action chunk predicted by the slow policy in an auto-regressive manner.}
    \label{fig:framework}
    \vspace{-5mm}
\end{figure*}

Previous works \cite{diffusion_policy,zhao2023learning} have demonstrated that predicting action sequences or action chunks \cite{lai2022action} effectively preserves temporal action consistency and handles non-Markovian or idle actions, which achieve superior performance in policy learning. However, when executing the action chunk, such approaches can be viewed as an open-loop policy, preventing it from achieving real-time feedback from high-frequency signals such as tactility. VISK \cite{pattabiraman2024learning} utilizes temporal ensembling to mitigate this issue. As shown in Fig. \ref{fig:pipeline_comparison}, temporal ensembling finds a balance between closed-loop control and sequence consistency by aggregating the predictions of multiple iterations for the same timestep. A significant drawback of this solution is that it diminishes the policy's ability to model multi-modal distributions and non-Markovian actions, making it prone to issues such as getting stuck. Moreover, we find that the policy performance is quite sensitive to the smoothing coefficient of temporal ensembling, which consequently reduces the policy's applicability.

To break the above trade-off between sequence modeling and closed-loop control, we propose a slow-fast policy learning framework \alg (RDP) as in Fig. \ref{fig:framework}. RDP is a slow-fast Latent Diffusion Model (LDM) \cite{rombach2022high}, which explicitly processes signals of various frequencies at different stages. In particular, we first convert the original action chunks to the latent space by training an asymmetric tokenizer (AT). The AT decoder takes the instantaneous tactile representation apart from the latent action chuck as input. For policy learning, a slow Latent Diffusion Policy (LDP) is trained to predict the latent action chuck according to the observation in a way similar to Diffusion Policy \cite{diffusion_policy}. During inference, we sample latent action chunks at a lower frequency, and within each chunk, actions are executed while the latest tactile representations are fed into the decoder of the AT to predict the real action for the next frame. This hierarchical design enables the slow network to maintain its capacity for modeling complex or non-Markovian actions by predicting temporally consistent latent action chunks and the fast network to achieve closed-loop control through real-time responsiveness. Next, we will discuss the design choices of each component within \alg.

\subsubsection{Fast Policy}
The fast asymmetric tokenizer (AT) consists of a 1D-CNN encoder $\mathscr{E}$ and a GRU \cite{cho2014properties} decoder $\mathscr{D}$. Given an action chunk $\mathbf{A}\in \mathbb{R} ^{T\times D}$ in the policy learning $\mathcal{D}_{policy}$, the encoder downsamples it to a latent one $\mathbf{Z} = \mathscr{E}(\mathbf{A})\in \mathbb{R}^{t\times d}$. We choose to use a CNN-based encoder to preserve the spatial structure of the raw sequence, enabling the latent action chunk to be better processed by the latent diffusion policy, which takes sequences as input. After that, the decoder reconstructs the action via $\hat{\mathbf{A}} = \mathscr{D}(\text{concat}([\mathbf{Z}, \mathbf{F}^{reduced}]))$, where $\mathbf{F}^{reduced}$ is the corresponding tactile representation sequence proposed in Sec. \ref{sec:tactile_representation}. It is worth noting that we utilize tactile representation solely as input in the decoder. This deliberate asymmetry in structure is designed to ensure that the latent action chunk retains only high-level feedback strategies, while the precise locations are predicted by the decoder with the tactile information. The AT is trained using an L1 reconstruction loss and a Kullback-Leibler (KL) penalty loss \cite{kingma2013auto} as in Eq. \ref{eq:loss_at}. 
\begin{equation}
    \label{eq:loss_at}
    L_{AT} = \mathbb{E}_{(\mathbf{A}, \mathbf{F}^{reduced})\in\mathcal{D}_{policy}} \left[\|\mathbf{A} - \hat{\mathbf{A}}\|_1 + \lambda_{KL}L_{KL}\right].
\end{equation}
In practice, we keep the coefficient $\lambda_{KL}$ small as in LDM \cite{rombach2022high} because we want to smooth the latent space of the AT rather than turning it into a generative model. As shown in Tab. \ref{tab:inference_time}, our fast policy only takes less than 1ms for inference, which can even support higher-frequency inputs ($>300$Hz) theoretically.

\subsubsection{Slow Policy}
We model the slow policy as a Diffusion Policy \cite{diffusion_policy} operating on latent action chunks, which is called Latent Diffusion Policy (LDP). Diffusion Policy is a generative model that iteratively denoises the noisy action $\mathbf{A}^k$ to a clean one $\hat{\mathbf{A}}^0$ through Stochastic Langevin Dynamics \cite{welling2011bayesian} with the learned gradient field $\nabla E(\mathbf{A})$. To transform the model to latent space, we use the latent action chunk $\mathbf{Z}^0 = \mathscr{E}(\mathbf{A}^0)$.
This modeling method offers several advantages. On the one hand, the downsampled latent representation reduces computational costs. More importantly, the asymmetric design in the AT allows challenging reactive behaviors to be excluded from latent action chunks, thereby reducing the learning difficulty of latent diffusion policy under low-frequency observation and enhancing its generalization capabilities.
During training, given the observation $\mathbf{O}$ (including image, tactility and proprioception), the gradient field is learned by $\epsilon_\theta$ and the DDPM training objective can be rewritten as
\begin{equation}
    L_{LDP} = \mathbb{E}_{(\mathbf{O},\mathbf{A}^0)\in\mathcal{D}_{policy}, k, \epsilon^k} \| \epsilon^k - \epsilon_{\theta}(\mathbf{O}, \mathbf{Z}^0  + \epsilon^k, k) \|_2 ,
\end{equation}
where $k$ is the iteration index and $\epsilon^k$ is a random noise with certain variance. We use CNN-base Diffusion Policy with FiLM-based \cite{perez2018film} condition injection as the network architecture.

\begin{table}[!ht]
    \centering    
    \caption{Inference Time of Different Modules on RTX 4090}
    \label{tab:inference_time}
    \resizebox{\linewidth}{!}{        
        \begin{tabular}{ccc}
            \toprule
             Diffusion Policy&  Slow Policy (LDP)& Fast Policy (AT)\\
             \midrule
             120ms&  100ms& $<1$ms\\
             \bottomrule
        \end{tabular}        
    }
    \vspace{-3mm}
\end{table}
\subsubsection{Implementing Suggestions for Slow-Fast Policy}
Compared to the standard Diffusion Policy \cite{diffusion_policy}, our slow-fast control policy requires certain key design elements to achieve optimal performance.
\begin{itemize}
    \item \textbf{Relative trajectory.} We use relative end-effector (EE) trajectory for action representation, which has been proven to be effective even in complex tasks by UMI \cite{chi2024universal}. Specifically, instead of directly calculating the delta action between consecutive frames (may lead to large compounding errors), we convert an absolute pose trajectory to a relative one by calculating the relative transformation with respect to a base frame. In our setting, the base frame is the last observation frame in an action chunk. 
    \item \textbf{Latency matching.} We calculate the latency caused by policy inference and action execution, and discard the first few action steps predicted by the model to send the accurately matched actions to the robot. This method has been mentioned in UMI \cite{chi2024universal} and is even more crucial for our slow-fast policy. It ensures smoother transitions between action chunks, preventing out-of-distribution tactile signals from causing the fast policy to predict abnormal actions.
\end{itemize}

\section{Experiments}
We design experiments to answer the following questions: 
\begin{itemize}
\item \textbf{Q1}: Does tactile signals improve policy performance in contact-rich tasks?
\item \textbf{Q2:} What are the capabilities of RDP?
\item \textbf{Q3:} Are RDP applicable to different tactile / force sensors?
    \item \textbf{Q4}: Can RDP react immediately to external perturbations? 
    \item \textbf{Q5:} Which design choices matter for RDP? Why can't we simply use small chunk size or temporal ensemble to increase closed-loop control frequency? 
    \item \textbf{Q6:} How does tactile / force feedback in \teleop contribute to the data quality in teleoperation?
    \item \textbf{Q7:} How does data quality in teleoperation influence policy performance?

\end{itemize}

\subsection{Setup}

\begin{figure*}[h]
    \centering
    \includegraphics[width=\linewidth]{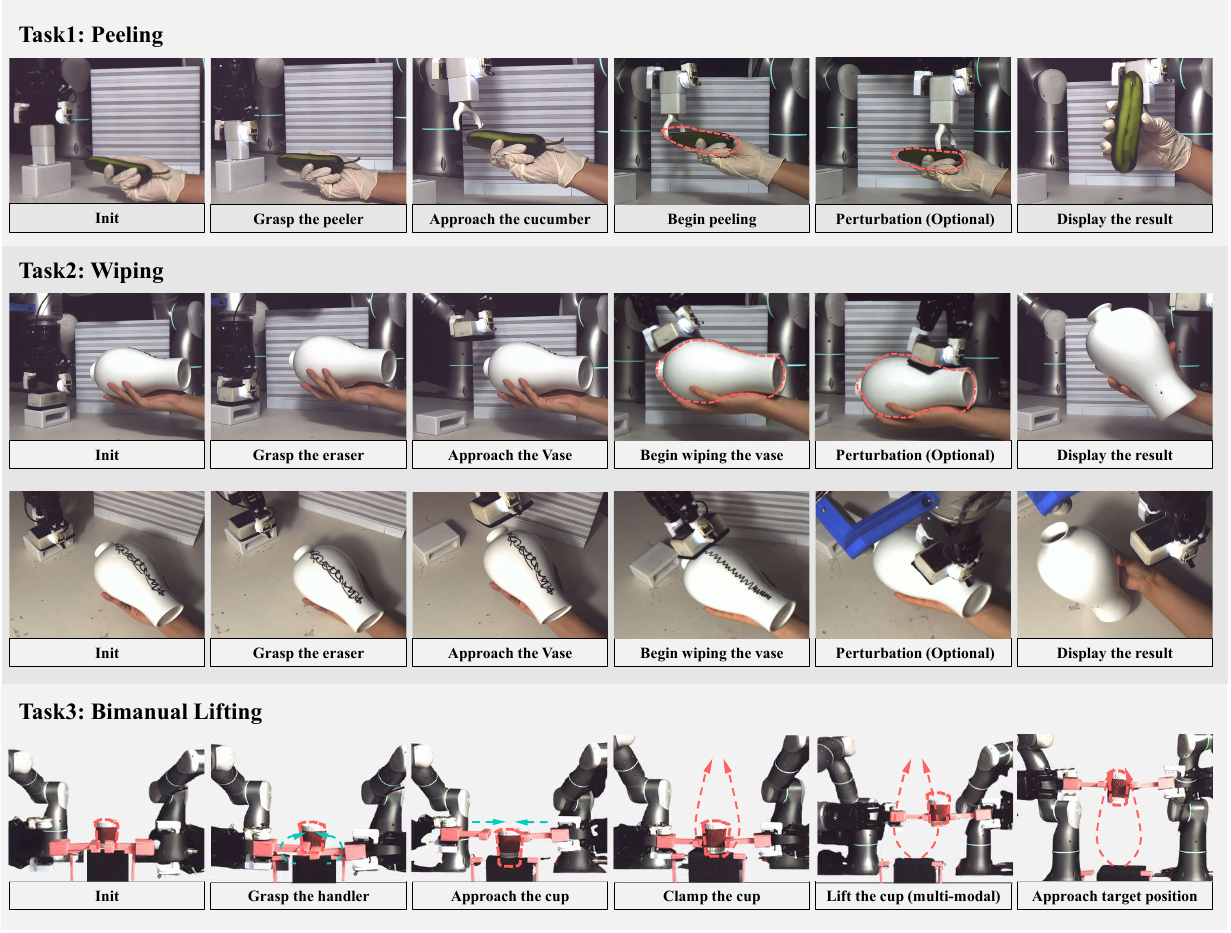}
    \caption{Three experiment tasks including \textit{Peeling}, \textit{Wiping} and \textit{Bimanual Lifting}.}
    \label{fig:task}
\vspace{-4mm}
\end{figure*}
\subsubsection{Hardware}
The experimental platform consists of two Flexiv Rizon 4 \cite{flexiv_rizon4} robotic arms with joint torque sensors and two Flexiv Grav \cite{flexiv_grav} grippers.  For single-arm tasks, we only use one Realsense D435 camera on the robot arm for the wrist view. For the bimanual task, we use two Realsense D435 cameras for wrist views and a fixed Realsense D415 camera in front of the robot workspace for external view. We use three different tactile / force sensors for experiments: 
\begin{itemize}
    \item \textbf{GelSight Mini} \cite{gelsight_mini} (Robotics Package) optical tactile sensor with 8MP resolution at 25 FPS, and it has a $7\times9$ marker dot array on the surface. 
    \item \textbf{MCTac} \cite{mctac} optical tactile sensor with 2MP resolution at 30 FPS, and it has a $5\times 7$ marker dot array on the surface. We have improved the original design of MCTac \cite{mctac}, including increasing the size of the marker, reducing the density of the marker, and using white lightning for better tracking stability of the marker.  Please see the Appendix \ref{sec:supp_hardware} for the hardware details of the improved MCTac sensor. 
    \item Built-in \textbf{joint torque sensors} in Flexiv Rizon 4 \cite{flexiv_rizon4} robotic arm. We use the estimated TCP force/torque calculated by Flexiv RDK \cite{flexiv_rdk} for experiments. We stream the sensor data at 120Hz and downsample it to 24 FPS. Note that the estimated TCP force / torque signals have relatively larger noise compared to the force sensor mounted on the robot end effector (\eg ATI mini 45\cite{ati_mini45}) due to inaccurate dynamics model, which further challenges the learning algorithm.
\end{itemize} In order to evaluate policy performance under different tactile / force sensors, we attach MCTac and GelSight Mini to different fingertips of the same gripper. In this way, we can collect synchronized data from MCTac, GelSight Mini and force/torque sensors simultaneously. The \teleop teleoperation uses a Meta Quest 3 VR headset. All devices are connected to a workstation with an Intel Core i9-14900K CPU and an NVIDIA RTX 4090 GPU for both data collection and evaluation.

\subsubsection{Baselines}
We use the following baselines for comparison:
\begin{itemize}
    \item \textbf{Diffusion Policy}: vanilla implementation of Diffusion Policy \cite{diffusion_policy} with only visual input (RGB images) and open-loop action chunking. 
    \item \textbf{Diffusion Policy (tactile image)}: Diffusion Policy with raw tactile images and visual input.
    \item \textbf{Diffusion Policy (tactile embedding)} Diffusion Policy with tactile embeddings (PCA feature) and visual input.
    \item \textbf{Reactive Diffusion Policy (tactile embedding) (Ours)}: our slow-fast policy with high-frequency tactile embedding (PCA feature) and visual input.
    \item \textbf{Reactive Diffusion Policy (force) (Ours)}: our slow-fast policy with high-frequency wrench (force/torque) and visual input.
\end{itemize}

\subsubsection{Tasks}
As shown in Fig. \ref{fig:task}, we evaluate \alg with three challenging contact-rich tasks. 
\begin{itemize}
    \item \textbf{Peeling.} The robot needs to grasp the peeler, approach a cucumber held midair by a human hand, then begin peeling. This task requires the following capabilities: (1) \textbf{Precision.} The robot needs to finish the task under environment uncertainties (\eg different tool grasp locations, different cucumber pose) with high precision (millimeter-level). (2) \textbf{Fast response}. The robot needs to react instantly to human perturbations. 
    \item \textbf{Wiping.} The robot needs to grasp the eraser, approach the vase held midair by a human hand, then begin wiping. This task requires the following capabilities: (1) \textbf{Adaptive force control with rotation.} The robot needs to adaptively track the curved vape surface with different environment uncertainties (\eg tool grasp locations, vase pose). (2) \textbf{Fast response}. The robot needs to react instantly to human perturbations. 
    \item \textbf{Bimanual Lifting.} The two robot arms need to grasp the handlers, approach the paper cup, clamp the paper cup with the two handlers, carefully lift the cup along the trajectory of the curve without squeezing it. This task requires the following capabilities: (1) \textbf{Precise force control.} The two robots must apply precise force during the task execution. It is crucial to avoid exerting excessive force that could squeeze the cup while also ensuring that the force is sufficient to prevent the cup from slipping. (2) \textbf{Bimanual coordination}. (3) \textbf{Multi-modality.} As shown in Fig. \ref{fig:task}, in the expert data, there are two upward lift trajectories, indicating the presence of multi-modality. 
\end{itemize}
\subsubsection{Evaluation Protocols}
We use \textit{similar initial states} across all methods for both the robots and the objects, by manually aligning the scene with the pre-defined images. 
There are three test-time variations for \textit{Peeling} and \textit{Wiping} tasks: (a) \textit{No perturbation.} The object is fixed with a random 6D pose in the air. (b) \textit{Perturbation before contact.} The human evaluator will move the object right before the tool makes contact. (c) \textit{Perturbation after contact.} The human evaluator will move the object after the tool makes contact to break the contact state. There are two test-time variations for \textit{Bimanual Lifting} task: (a) \textit{soft paper cup.} (b) \textit{hard paper cup.} We run 10 trials for each test-time variation. 

For \textit{Peeling} task, we calculate the score based on the proportion of the peeled cucumber skin to the total length of the cucumber, normalized by the average score of the demonstration data. For \textit{Wiping} task, we calculate the score based on the size of the remaining handwriting compared to the demonstration data. If the residue reaches the human demonstration level, the score is 1; If there is minor residue (less than one third of the handwriting length), the score is 0.5; If significant residue remains, the score is 0. For \textit{Bimanual Lifting} task, if the paper cup is lifted into the air following the designated trajectory without significant compression, the score will be 1; If the paper cup is partially compressed in the air, the score will be 0.5; If the cup is not lifted up, or dropped in the air, the score will be 0. Please see Appendix \ref{sec:supp_evaluation} for more details of the evaluation protocal.

\subsubsection{Implementation Details}
The Diffusion Policy and our slow policy (LDP) predict open-loop 12 FPS action sequences for each action chunk. The low-frequency observation of LDP includes both visual inputs and tactile / force inputs. The fast policy (AT) takes tactile / force observations at 24 FPS and update new action predictions at 24 FPS.  Note that we use 24 FPS because we are constrained by the frame rate limitation of GelSight \cite{gelsight_mini}, which is 25 FPS. Our RDP algorithm can also be applied to higher frequency tactile / force signals in theory. Please see Appendix \ref{sec:supp_dataset}, \ref{sec:supp_impl_details} and \ref{sec:supp_hyperparam} for more details on data collection, the inference process and the hyperparameters.

\subsection{Results}
\textbf{Simply adding tactile signals into observation may NOT improve performance for complex contact-rich tasks (Q1).}
We have compared the performance of Diffusion Policy using raw tactile images (DP w. tactile img.) v.s. low-dim tactile embedding (DP w. tactile emb.) in Tab. \ref{tab:peeling_results}. Although the performance of both methods is similar, low-dimensional tactile embedding demonstrates greater robustness to texture changes resulting from gel damage or gel replacements during the evaluation process. Therefore, we use tactile embedding in most of our experiments.

\begin{figure*}[!ht]
    \centering
    \includegraphics[width=\linewidth]{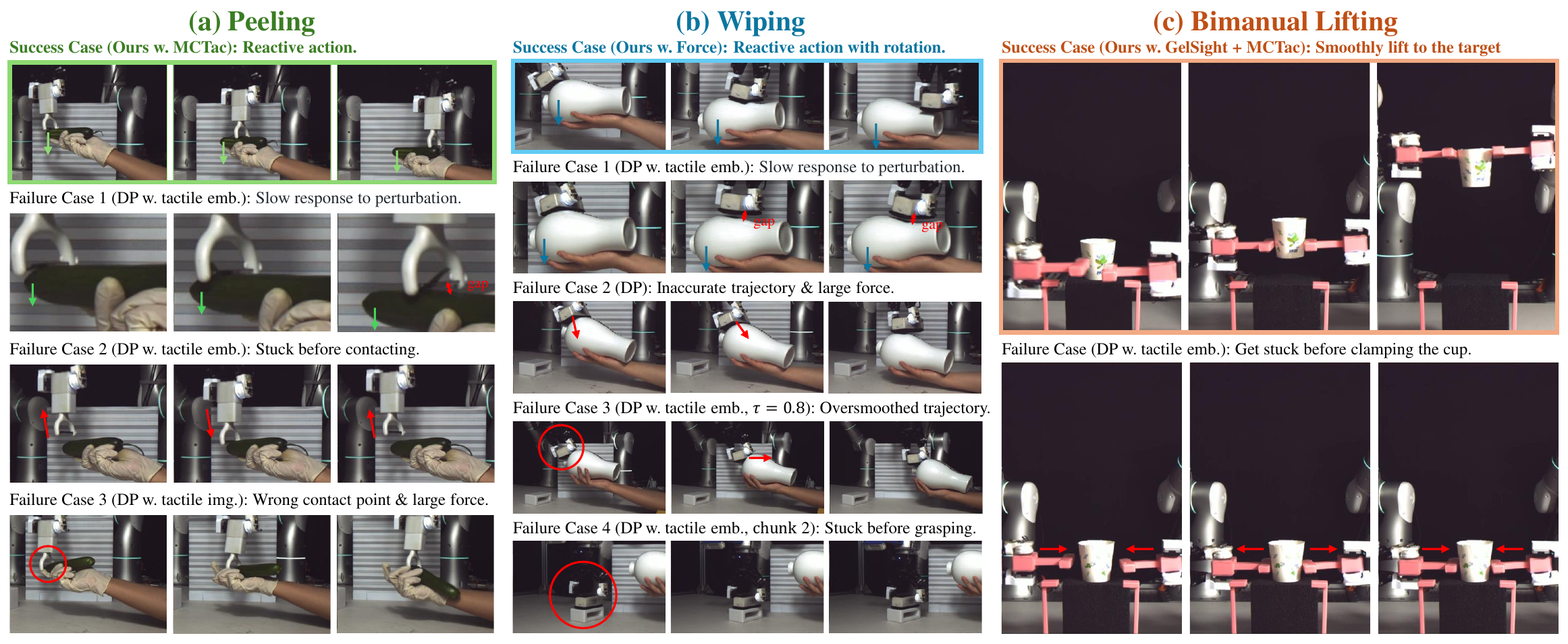}
    \caption{Evaluation results and failure cases of baselines. Please see the \href{https://reactive-diffusion-policy.github.io/}{website} for more details.}
    \label{fig:main_results}
    \vspace{-4mm}
\end{figure*}

We also find that Diffusion Policy (DP) incorporates tactile embedding (DP w. tactile emb.) performs similarly compared to DP with purely visual inputs in the three tasks (see Tab \ref{tab:peeling_results}, Tab \ref{tab:wiping_results} and Tab \ref{tab:lifting_results}). However, despite similar performance, these two DP baselines exhibit different failure modes. We observe that DP with pure visual input frequently predicts inaccurate trajectories and results in large contact forces (\eg failure case 2 in Fig. \ref{fig:main_results} (b)), necessitating the human evaluators to move their hands to prevent sensor damage. In contrast, DP with tactile embedding rarely brings large contact forces. It may get stuck when making contact with the object (\eg failure case 2 in Fig. \ref{fig:main_results} (a) \& failure case in Fig. \ref{fig:main_results} (c)). This may be because it learns some reactive behavior from the data (\eg slightly move up when making contact; move down when loose contact). However, since it executes action chunks in an open-loop manner, it lacks the capability for finer adjustments. Consequently, it might quickly transition to a different contact state and repeatedly switch between various contact states. In conclusion, we need a better way to incorporate tactile signals into the policy.

\begin{table}[!ht]
    \centering
\caption{Policy Performance for Peeling Task}
\label{tab:peeling_results}
\resizebox{\linewidth}{!}{
    \begin{tabular}{l|ccc|c}
        \toprule
         & No& Perturb.&Perturb.& All \\
 & Perturb.& before Contact& after Contact&\\
        \midrule
        DP& 0.56 & 0.58 & 0.19 & 0.44\\
        DP w. tactile img.& 0.60& 0.49& 0.16& 0.41\\
 DP w. tactile emb.& 0.48 & 0.55 & 0.15 & 0.39\\
        \midrule
        RDP (GelSight)& 0.98 & 0.93 & 0.80 & 0.90 \\        
        RDP (MCTac)& \textbf{1.00}& 0.84 & 0.79 & 0.88 \\
        RDP (Force)& 0.99 & \textbf{0.98} & \textbf{0.88} & \textbf{0.95} \\
 \bottomrule
    \end{tabular}
    }        
    \vspace{-3mm}
\end{table}

\begin{table}[!ht]
    \centering
\caption{Policy Performance for Wiping Task}
\label{tab:wiping_results}
\resizebox{\linewidth}{!}{
    \begin{tabular}{l|ccc|c}
        \toprule
         & No& Perturb.&Perturb.& All \\
 & Perturb.& before Contact& after Contact&\\
        \midrule
        DP& 0.75& 0.70& 0.25 & 0.57\\
        DP w. tactile emb.& 0.60& 0.75&0.15& 0.50\\
        \midrule
        RDP (GelSight)& 0.85& \textbf{0.95}& 0.50 & 0.77\\
        RDP (Force)& \textbf{0.95}& 0.85& \textbf{0.80}&\textbf{0.87}\\
        \bottomrule
    \end{tabular}
    }        
    \vspace{-4mm}
\end{table}

\begin{table}[!ht]
    \centering
\caption{Policy Performance for Bimanual Lifting Task}
\label{tab:lifting_results}
\resizebox{\linewidth}{!}{
    \begin{tabular}{l|ccc|ccc|c}
        \toprule
          & \multicolumn{3}{c|}{Soft Paper Cup} & \multicolumn{3}{c|}{Hard Paper Cup} & All\\
        \cmidrule(lr){2-4}
        \cmidrule(lr){5-7}
        \cmidrule(lr){8-8}
         & Clamp & Lift & Score & Clamp & Lift & Score & Score \\
        \midrule
        DP& $0\%$ & $0\%$ & 0.00 & $0\%$ & $0\%$ & 0.00 & 0.00 \\
        DP w. tactile emb.& 10\% & 10\% & 0.10 & $20\%$ & $10\%$ & 0.05& 0.08\\
        \midrule
        RDP (GelSight + MCTac)& $\mathbf{100\%}$ & $\mathbf{100\%}$& 0.55& $\mathbf{90\%}$ & $80\%$  & 0.40& 0.48\\
        RDP (Force)& $\mathbf{100\%}$ & $90\%$  & $\mathbf{0.80}$& $\mathbf{90\%}$ & $\mathbf{90\%}$ & $\mathbf{0.60}$ & $\mathbf{0.70}$\\
        \bottomrule
    \end{tabular}
    }        
    \vspace{-3mm}
\end{table}
\textbf{RDP can perform challenging contact-rich tasks that require precision, adaptive and precise force control, or bimanual coordination (Q2).}
As shown in Tab. \ref{tab:peeling_results}, Tab. \ref{tab:wiping_results} and Tab. \ref{tab:lifting_results},
RDP improves the overall score by a large margin ($>35\%$) compared to various Diffusion Policy baselines in all three tasks. These tasks require different capabilities, including precision (\textit{Peeling}), adaptive force control with rotation (\textit{Wiping}) and precise force control with bimanual coordination (\textit{Bimanual Lifting}). We believe that these capabilities are highly related to closed-loop adjustments with high-frequency tactile / force feedback. 

\begin{figure*}[!ht]
    \centering
    \includegraphics[width=\linewidth]{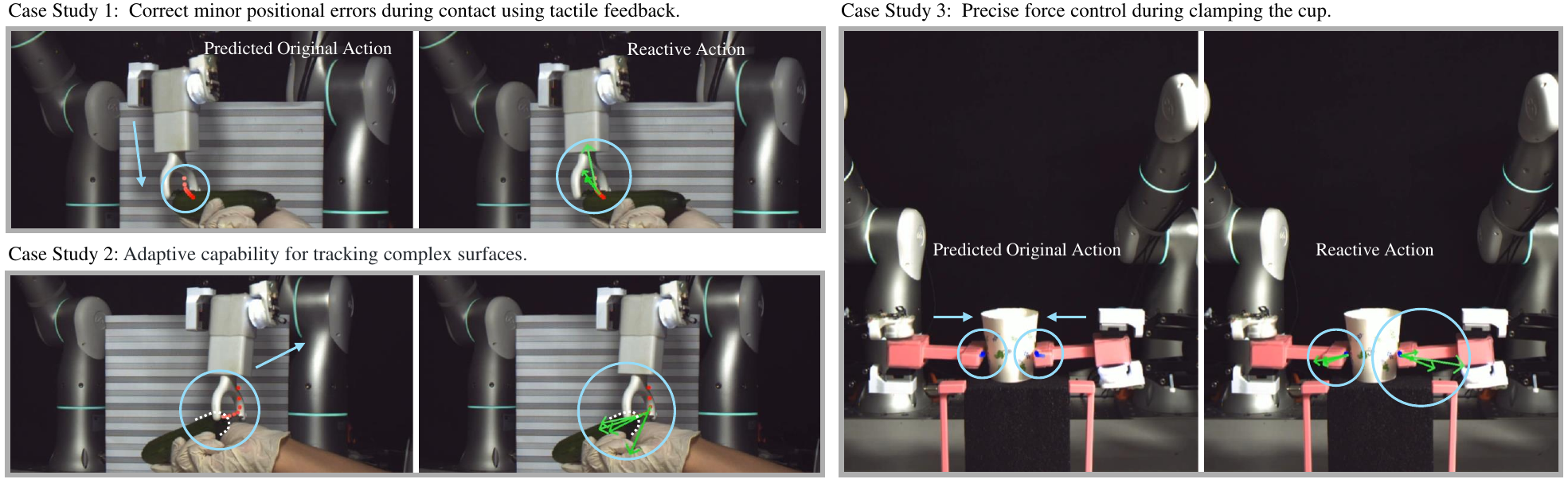}
    \caption{Visualization of the RDP inference process. The red (left) and blue (right) dots can be seen as the predicted action chunk of the slow policy.
    The green arrow represents the correction direction and magnitude (scaled up for better visibility) of the reactive action predicted by the fast policy during inference. Please see the videos on the \href{https://reactive-diffusion-policy.github.io/}{website} for more details.}
    \label{fig:policy_explain}
\vspace{-4mm}
\end{figure*}

We have performed some case studies and visualization to analyze how RDP works during these contact-rich tasks in Fig. \ref{fig:policy_explain}.   We have observed that RDP indeed learns reactive behaviors similar to those of humans. For instance, in Case Study 1, when a peeler suddenly comes into contact with a cucumber, the contact force increases abruptly. The fast policy then applies an upward adjustment to the action chunk predicted by the slow policy to reduce the contact force. In Case Study 2, as the peeler approaches the end of the cucumber, the fast policy predicts a downward corrective action, allowing the peeler to closely follow the cucumber's surface and remove more skin. In Case Study 3, when two handlers attempt to clamp the paper cup, the fast policy swiftly predicts an outward corrective action once contact is made, preventing the cup from being squeezed. The magnitude of these reactive behavior adjustments is at the sub-millimeter level, which might appear small but significantly impacts the overall performance of these contact-rich tasks. These minor reactive actions are hard to learn accurately for a policy with open-loop action chunking, but much easier to learn for our fast policy with high-frequency tactile / force control.

Particularly, in \textit{Bimanual Lifting} task, we have observed the multi-modal behavior of the RDP model with force input, which indicates that our slow policy (LDP) remains the powerful modeling capabilities of diffusion model. 

\textbf{RDP are applicable to different tactile / force sensors (Q3).}
As shown in Tab. \ref{tab:peeling_results},  the performance of RDP with the two optical tactile sensors (GelSight Mini \cite{gelsight_mini} and MCTac \cite{mctac}) are very close (0.9 vs 0.88) on \textit{Peeling} task. Furthermore, we surprisingly find that RDP can perform well using different tactile sensors (MCTac on the left gripper and GelSight on the right gripper) at the same time for the \textit{Bimanual Lifting} task (see Tab. \ref{tab:lifting_results}). Note that GelSight Mini and MCTac have different LED colors (RGB vs white), different gel material (hard vs soft), different marker arrays ($7\times 9$ vs $5\times 7$), different resolutions (8MP vs 2MP) and different frame rates (25 FPS and 30 FPS). This validates the effectiveness of our tactile representation and the versatility of the RDP algorithm. 

RDP is also applicable for force / torque sensors. Tab. \ref{tab:peeling_results}, Tab. \ref{tab:wiping_results} and Tab. \ref{tab:lifting_results} have shown that RDP with simple force input without any additional design achieves the best results in all three tasks. Despite the significant noise associated with the force sensor during rapid robot movements (as evidenced in the TactAR part of the supplementary video), the RDP algorithm successfully identifies useful patterns from the noisy data. We hypothesize that the force sensor may yield better results due to its lower latency and reduced dimensionality compared to optical tactile sensors, which potentially facilitates easier learning by the network. 

\textbf{RDP can react immediately to external disturbances (Q4).}
We find that the slow-fast design of RDP significantly improved the model's response speed to external disturbances. For example, in Tab. \ref{tab:peeling_results}, RDP (GelSight) achieves a score of 0.8 on the hardest setting (Perturbation after Contact) in \textit{Peeling} task, while the DP baseline with tactile embedding only achieves a score of 0.15. In the \textit{Wiping} task (see Tab. \ref{tab:wiping_results}), we find that RDP also achieves a higher score compared to DP baselines under human perturbations. Fig. \ref{fig:main_results} has shown some typical failure cases for DP baselines under human perturbations (\eg failure case 1 in \textit{Peeling} (a) task and \textit{Wiping} (b) task). When the robot loses contact with perturbations (\eg moving down), the DP baselines will execute the remaining trajectory in the action chunk in an open-loop manner, resulting in broken cucumber peels and residual handwriting on the vase that were not completely wiped off. In contrast, the RDP algorithm can immediately change the predicted trajectory with the fast policy in a closed-loop manner, which leads to better results.

\textbf{Slow-fast hierarchy, relative trajectory and latency matching are essential for RDP performance (Q5).}
As shown in Fig. \ref{fig:pipeline_comparison}, there are two ways to increase the closed-loop control frequency without our slow-fast hierarchy: (1) reducing action chunk size. (2) using temporal ensemble. However, experiments in Tab. \ref{tab:temporal_ensemble} have proved that these two options both have significant side effects. We can see from Tab. \ref{tab:temporal_ensemble} that when the action chunk size is reduced from 8 to 2, the DP baseline tends to get stuck before grasping (failure case 4 in Fig. \ref{fig:main_results} (b)), which makes the grasp success rate drop from $100\%$ to $20\%$. Policy with small chunk size is very sensitive to non-Markovian behaviors (\eg pauses in the air) commonly found in human demonstration data, so we can not simply reduce chunk size. 

\begin{figure}[!ht]
    \centering
    \includegraphics[width=\linewidth]{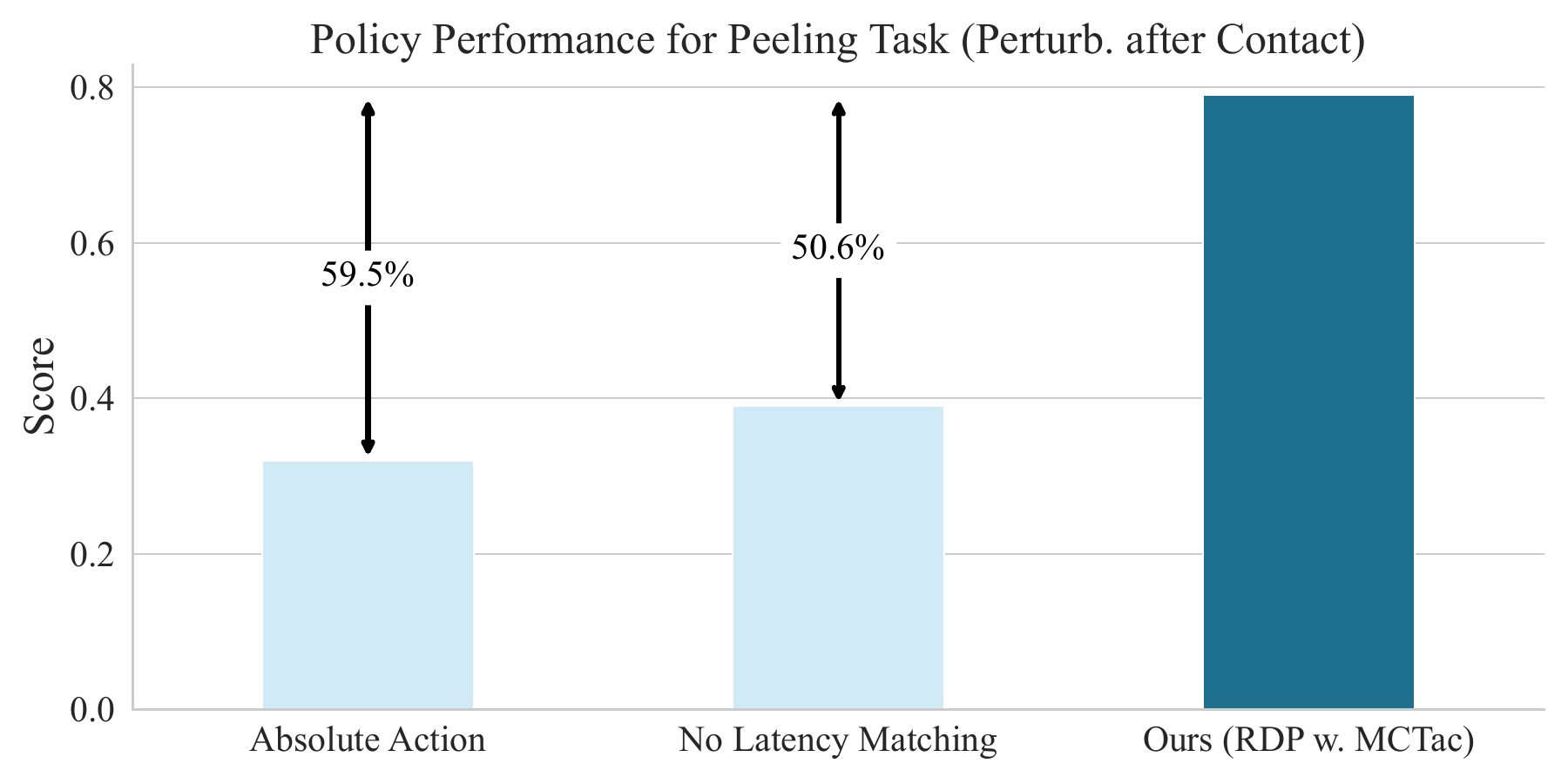}
    \caption{Ablation Study.}
    \label{fig:ablation}
    \vspace{-3mm}
\end{figure}
\begin{table}[!ht]
    \centering
\caption{Effects of Chunk Size and Temporal Ensemble}
\label{tab:temporal_ensemble}
\resizebox{\linewidth}{!}{
    \begin{tabular}{lcc|cc}
    \toprule
  & && \multicolumn{2}{c}{Wiping}\\        
   \midrule
          &  action chunk &temporal ensemble \cite{hato} & \multicolumn{2}{c}{Perturb. after Contact}\\
          \cline{4-5}
  & size&factor  & Grasp & Score\\
 \midrule
 \multirow{2}{*}{DP w. tactile emb.}&  8&- & $\mathbf{100}\%$ & 0.15\\
 & 2& - & $20\%$ & 0.10 \\
        \midrule
         \multirow{3}{*}{DP w. tactile emb.}&  8&$\tau$=0.2 & $30\%$ & 0.05 \\
 & 8&$\tau$=0.5& $0\%$ & 0.00 \\
        & 8&$\tau$=0.8& $\mathbf{100}\%$ & 0.15 \\
    \midrule
 RDP (GelSight)& 8& - & $\mathbf{100}\%$ & $\mathbf{0.50}$\\
 \bottomrule
    \end{tabular}
    }        
    \vspace{-3mm}
\end{table}

Temporal ensemble \cite{act, hato} can perform semi-closed-loop control by averaging predictions from multiple timesteps. We have also tried different temporal ensemble factors $\tau$ in HATO \cite{hato}, and the experiments in Tab. \ref{tab:temporal_ensemble} have shown that the model performance is very sensitive to $\tau$. When $\tau=0.2$, the average weight will focus more on the newest predictions, which makes the model behavior similar to the policy with small chunk size and causes low grasp rate ($30\%$). When $\tau=0.8$, the average weight will focus more on the oldest predictions, which makes the model behavior over-smoothed and hurt reactive ability (failure case 3 in Fig. \ref{fig:main_results} (b)). Thus, it is very hard to balance temporal consistency and reactivity with temporal ensemble.

As shown in Fig. \ref{fig:ablation}, the relative trajectory prediction performs much better compared to the absolute action prediction in \textit{Peeling} Task.  It may be because relative trajectory are easier to learn for a smaller, fast policy, which brings a more generalizable reactive strategy from tactile feedback. In addition, the relative trajectory also compresses the latent space, facilitating the learning process of the latent diffusion policy. We also find that latency matching also contributes a lot to the policy performance (see Fig. \ref{fig:ablation}) by ensuring smooth action transition between action chunks and reducing out-of-distribution (OOD) behaviors. 

Please see Appendix~\ref{sec:supp_more_ablation_results} for more ablation results.

\begin{figure}[!ht]
    \centering
    \includegraphics[width=\linewidth]{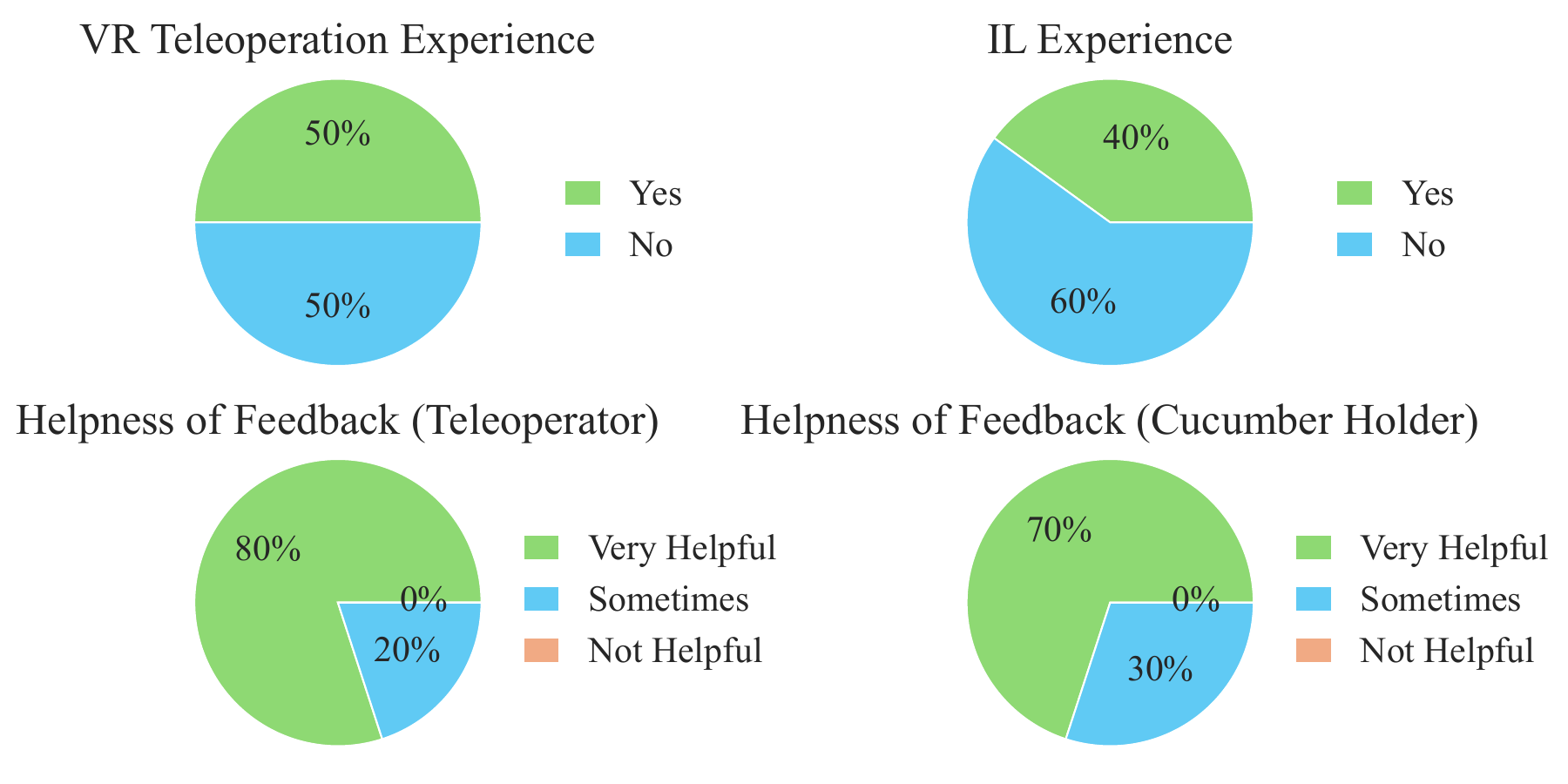}
    \caption{User study results among 10 users on teleoperation w./w.o. tactile / force feedback in \textit{Peeling} task.}
    \label{fig:tactar_user_study}
    \vspace{-3mm}
\end{figure}

\begin{figure}[!ht]
    \centering
    \includegraphics[width=\linewidth]{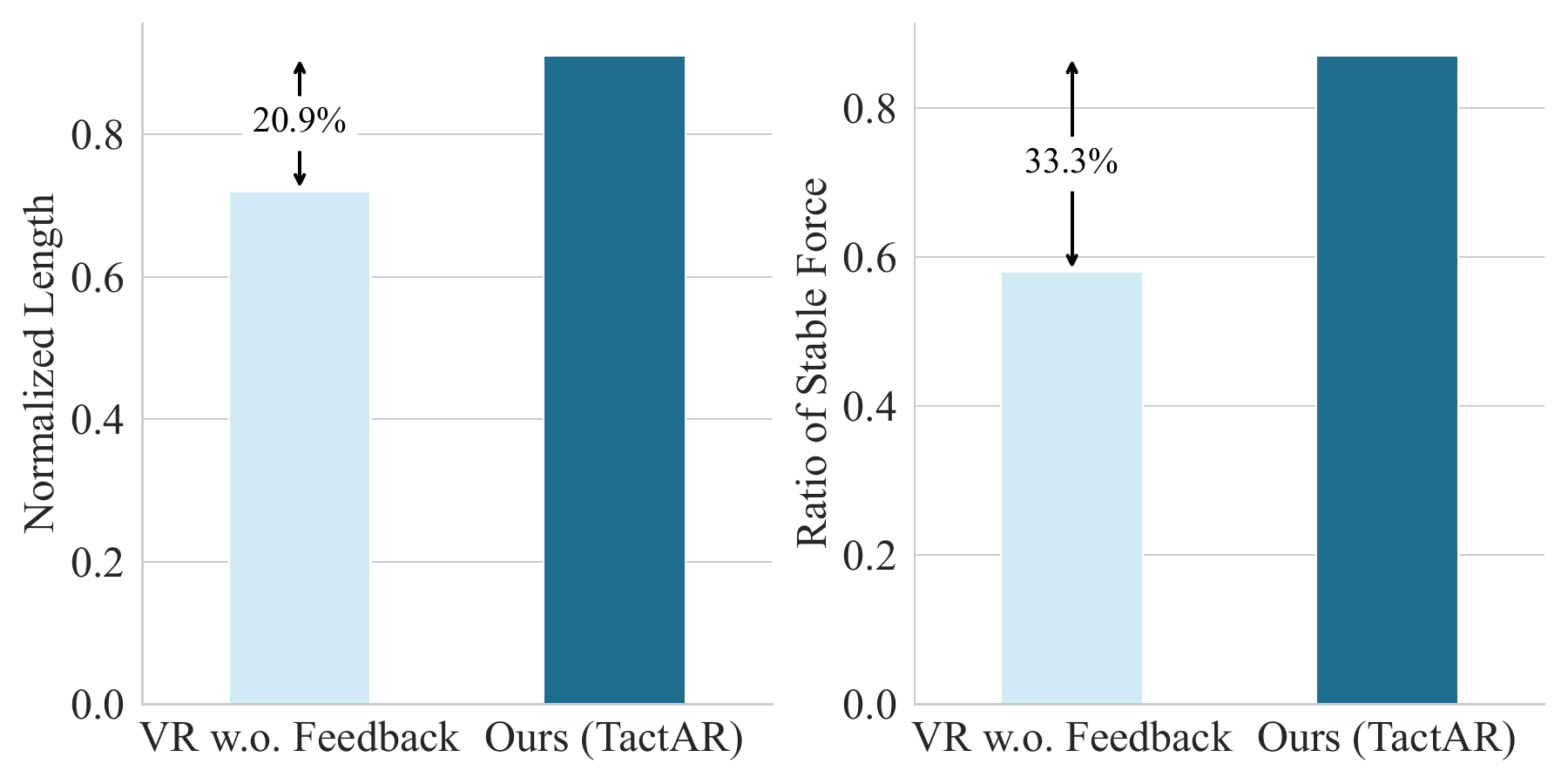}
    \caption{Teleoperation data quality of 10 users on \textit{Peeling} task (no perturb.) w./w.o. tactile / force feedback.}
    \label{fig:teleop_data_quality}
    \vspace{-5mm}
\end{figure}

\textbf{Tactile / force feedback in TactAR improves data quality in contact-rich tasks by improving the stability of contact forces (Q6).}
We have conducted a user study on how tactile / force feedback in TactAR helps the data collection process. We invited 10 users with different levels of experience in VR teleoperation or Imitation Learning (IL). Please see Appendix \ref{sec:supp_user_study_details} for more details of the user study.  We perform $10\times10\times2=200$ trials in total for quantitative analysis. As shown in Fig. \ref{fig:tactar_user_study}, most of the users ($\ge70\%$) found that tactile / force AR feedback is very helpful in data collection, regardless of whether they were acting as the teleoperator or the person holding the cucumber. In Fig. \ref{fig:teleop_data_quality}, we can see that tactile / force feedback in \teleop can greatly improve the data quality from both the normalized peeling length ($0.72\rightarrow 0.91$) and the ratio of stable contact force ($0.58\rightarrow 0.87$).

\begin{figure}[!ht]
    \centering
    \includegraphics[width=\linewidth]{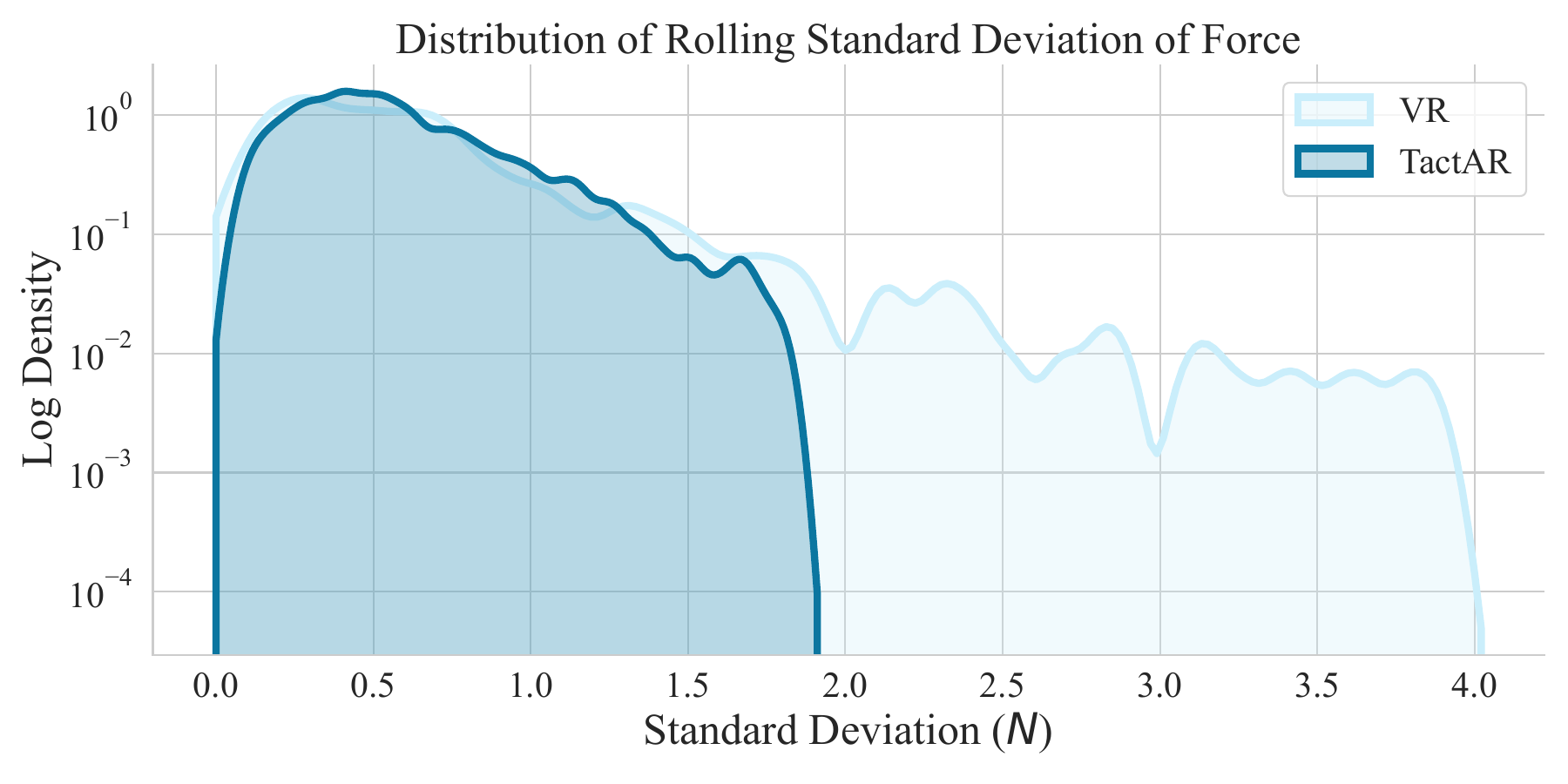}
    \caption{The stability of contact forces with different teleoperation systems in \textit{Peeling} task. Data collected with TactAR has higher stability of contact forces compared to traditional VR teleoperation.}
    \label{fig:data_quality}
    \vspace{-5mm}
\end{figure}

For a more detailed quantitative analysis of contact forces, we collect data (10 demos) for \textit{Peeling} task with the same user by VR teleoperation and TactAR respectively, then we calculate the Rolling Standard Deviation of the recorded force curve with a window size of 10 steps, and the results are shown in Fig. \ref{fig:data_quality}. We can observe that using \teleop to collect data helps avoid a large rolling standard deviation, indicating reduced temporal fluctuations and more stable contact forces.  

\begin{figure}[!ht]
    \centering
    \includegraphics[width=\linewidth]{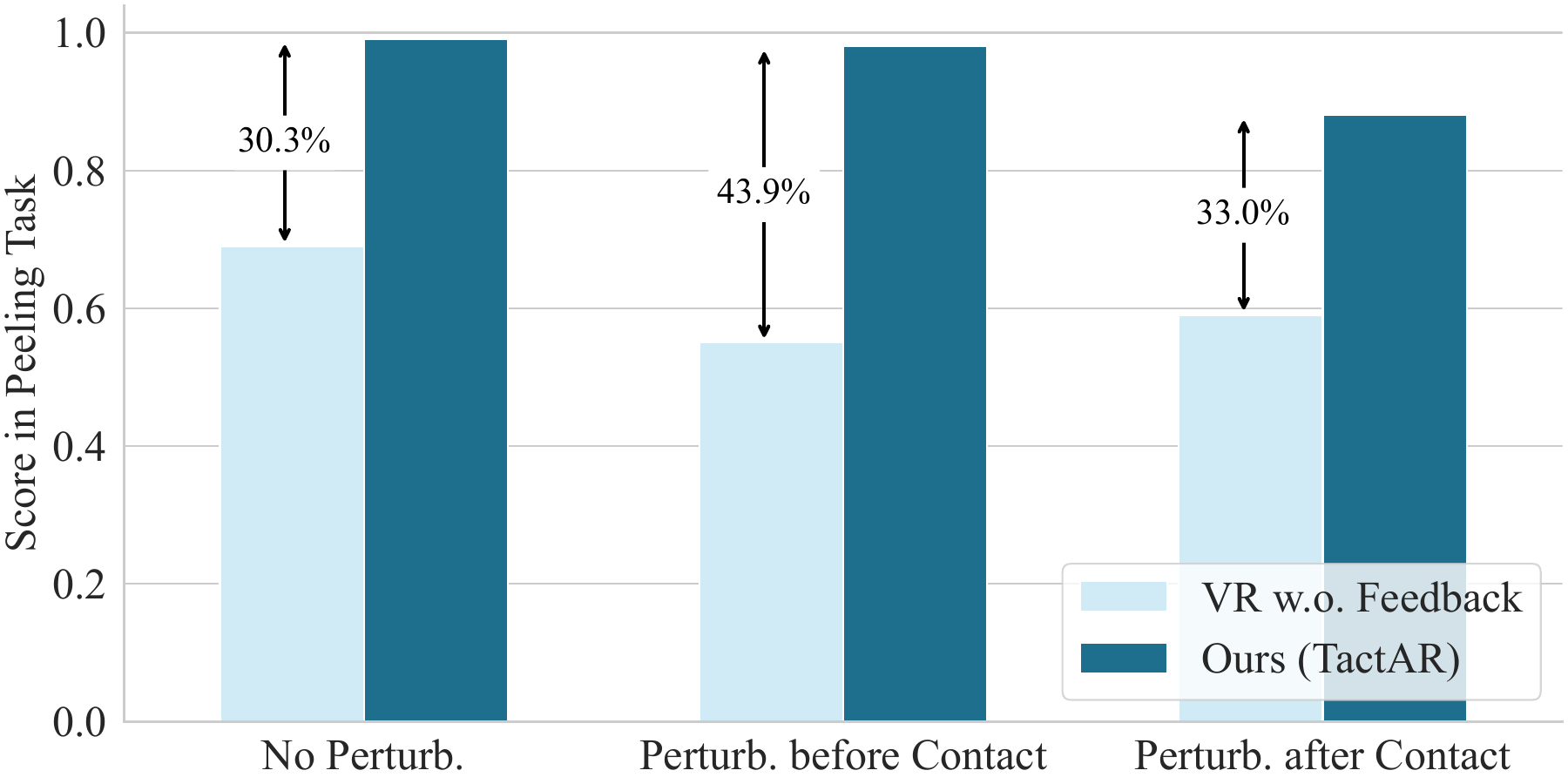}
    \caption{How data quality influences policy performance of RDP (force) in \textit{Peeling} task.}
    \label{fig:policy_performance_with_different_teleop}
    \vspace{-3mm}
\end{figure}

\textbf{High-quality data will help the model to discover useful patterns more easily (Q7).}
We have collected the same number of demonstrations (60) for \textit{Peeling} task with traditional VR teleoperation without tactile / force feedback and trained RDP (force) with these data. The results in Fig. \ref{fig:policy_performance_with_different_teleop} show that data quality has a large influence on policy performance (the score decreases by more than $30\%$). We observe that policies trained with low-quality data exhibited more unstable performance, such as more unstable force during peeling and a higher likelihood of breaking halfway through. A possible explanation is that the Fast Policy in RDP is designed to identify associations between tactile / force signals and trajectories from the data and learn reactive behavior. When contact forces in the data are highly unstable, the Fast Policy struggles to identify reasonable associations, which reduces performance.

\section{Limitations and Future Works}
While \teleop and RDP have demonstrated efficacy in numerous challenging tasks, they still have certain limitations. First, although \teleop can provide some degree of tactile / force feedback in AR, it is not as intuitive or efficient as direct human-hand operations. Future work could improve teleoperation efficiency by further reducing sensor and system latency. Second, our \teleop system is designed for two-finger grippers. Expanding our \teleop system and RDP algorithm to dexterous hands equipped with tactile sensors presents a promising direction for future research. Third, the fast policy in the RDP algorithm is currently limited to responding to high-frequency tactile / force input signals but cannot swiftly process high-frequency image inputs. Future research could consider incorporating high-frequency visual inputs into the fast network, enabling applicability to a broader range of task types. Lastly, the RDP algorithm is currently limited to single-task scenarios. Future work could integrate RDP with the Vision-Language Action (VLA) model by replacing the VLA tokenizer with an asymmetric tokenizer similar to the one in the RDP algorithm. This integration could introduce reactive behavior for closed-loop control with real-time tactile / force feedback within the VLA.
\section{Conclusion} 
\label{sec:conclusion}

In this paper, we present TactAR, a low-cost teleoperation system that provides real-time tactile / force feedback through AR, and Reactive Diffusion Policy (RDP), a novel slow-fast imitation learning algorithm for contact-rich manipulation. TactAR demonstrates that high-quality tactile / force feedback can be achieved through AR visualization without expensive specialized hardware. RDP successfully addresses the trade-off between sequence modeling and closed-loop control through its hierarchical design - using a slow network for complex trajectory planning and a fast network for reactive tactile feedback control. Through extensive experiments on three challenging contact-rich tasks, we demonstrated that RDP significantly outperforms state-of-the-art visual IL baselines in terms of both task completion and reactivity to tactile feedback. The cross-sensor experiments further validated that our approach generalizes well across different tactile / force sensors. We believe that this work takes an important step toward making visual-tactile imitation learning more practical and accessible. 

\section*{Acknowledgments}
This work is supported by the Shanghai Commitee of Science and Technology, China(Grant No.24511103200)
by the National Key Research,  Development Project of China (No.  2022ZD0160102), XPLORER PRIZE grants and Tsinghua Dushi program.

We thank Yongkai Fan for providing inspiration for the hardware design in our experiments. We thank Yi Wang, Junjie Fang, Fangyuan Zhou, Yuan Fang, Yanbing Zhou, Zimo Wen, Yutong Li, Wei Jiang, Yijin Chen and Yongkai Fan for participating in the user study. We also thank Hongjie Fang and Zihao He for providing help and feedback on manipulation with force sensing.


\bibliographystyle{plainnat}
\bibliography{references}

\clearpage
\appendix

\subsection{Hardware Details}\label{sec:supp_hardware}
\begin{figure}[!ht]
    \centering
    \includegraphics[width=\columnwidth,height=.5\columnwidth]{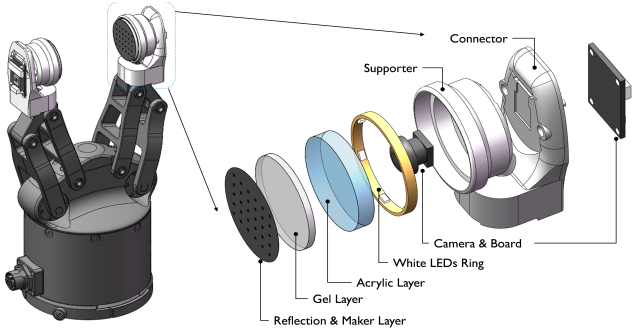}
    \caption{Improved MCTac Sensor for our task. The left part is the gripper integrated illustration, and the right part is the detailed structure and components of the camera-based tactile sensor.}
    \label{fig:markerTac}
\end{figure}

To acquire a stable and sensitive tactile signal, we have improved the camera-based tactile sensor based on the open source MCTac~\cite{mctac,mctac2}, details are shown in Fig. \ref{fig:markerTac}.
First, we redesign the mechanical interface module to fit the connector of Flexiv Grav Gripper \cite{flexiv_grav}. We also design a mount for the camera for more convenient cable management. 
Second, because we pay more attention to the shear force in this work, we utilized the white side illumination in stand of the colorful LEDs, which not only highlight the markers and their displacement, but also simplified the system complexity.
Finally, to increase the tracking stability and to avoid the tracking confusion (which can lead to abnormal values under large shear displacement), we have enlarged the gel marker size (1 mm) and their spacing (3 mm). This sparse marker array not only satisfied the tactile requirement, but also avoid the confusion and losing in marker tracking. After painting the reflection layer, we take the laser marking machine to process the surface with $5\times 6$ circle markers array (can remove the reflection layer in these regions), then we spray black ink on the proceeded surface to increase the contrast of tactile image, as shown in~\cite{taylor2022gelslim}.
After fabricating, we have assembled the customized camera-based tactile sensor, integrated it on the robot gripper, and token the USB-HUB to collect the observed tactile image stream.
\begin{figure}[!ht]
    \centering
    \includegraphics[width=0.7\linewidth]{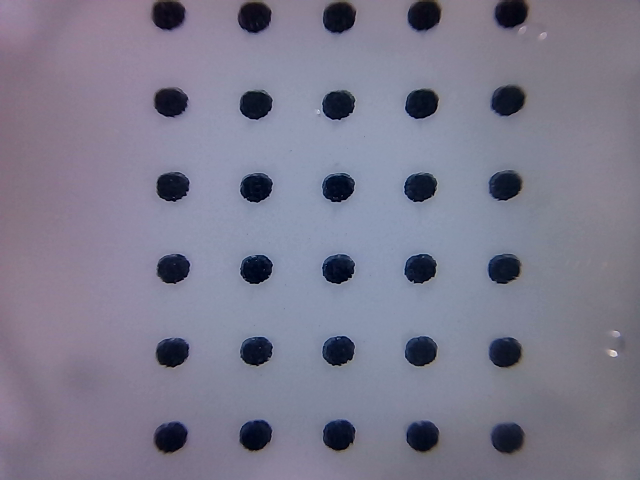}
    \caption{The example image of our improved MCTac optical tactile sensor.}
    \label{fig:mctac_image}
\end{figure}

\subsection{Implementation Details of TactAR}\label{sec:supp_tactar_details}
The VR headset will receive the latest robot TCP pose transformation matrix $W^{TCP}_t$, the 3D deformation field $V_t$ and the undeformed marker locations $D_0$. If we use multiple 3D arrows to represent the 3D deformation field in robot TCP coordinate system, we can calculate the start points $P_{start}^{TCP}$and the end points $P_{end}^{TCP}$ of the 3D arrows by Eq. \ref{eq:arrows_tcp}. Then, we convert the 3D arrows from the robot TCP coordinate system to the world coordinate system in AR by Eq. \ref{eq:arrows_ar}, here $s$ is the scale factor for better visibility.
\begin{equation}
    P_{start}^{TCP}=D_0,  P_{end}^{TCP}=D_0+V_t
    \label{eq:arrows_tcp}
\end{equation}
\begin{equation}
    P_{start}^{AR}=s W_t^{TCP} D_0,  P_{end}^{AR}=s W_t^{TCP} (D_0+V_t)
    \label{eq:arrows_ar}
\end{equation}
Finally, we render the 3D arrows $P_{start}^{AR}, P_{end}^{AR}$ in AR with Quest3. We adaptively change the color of a 3D arrow based on its length to enhance its visibility. As shown in Fig. \ref{fig:deformation_field}, the tangential force and the torsional torque can be easy to recognize from the 2D flow field, but the normal force that is typically manifested as an outward diffusion pattern can be challenging to observe intuitively. When the gripper is closed with large normal forces, the marker pattern may be dominant by the outward diffusion pattern. So we additionally implement an auto-detection mechanism, which automatically resets the visualized marker flow when the gripper is closed or open. 

Our system can achieve low-latency feedback for tactile/force sensors. Typically, the latency of the marker flow tracking algorithm is about 10ms. The rendering latency in Quest 3 is about 10ms. And the network latency is about 1ms - 6ms. In practice, the AR feedbacks of force sensors are more intuitive than optical tactile sensors, because force sensors has lower internal latency ($<1$ms) while optical tactile sensors have 10ms - 60ms internal latency depending on the exposure time, USB bandwith and image resolution.

\subsection{Details of the Tactile Representation}\label{sec:supp_tactile_pca}

For an optical tactile sensor with $n$ marker dots, we can calculate the 2D deformation field matrix $F$ with shape $(n, 2)$ according to Eq. \ref{eq:flow_2d} for each frame.  The deformation field is more robust and less affected by noise such as differences in lighting and texture when the gel on the sensor surface is replaced or slightly damaged during evaluation. 

\begin{figure}[h]
    \centering
    \includegraphics[width=\linewidth]{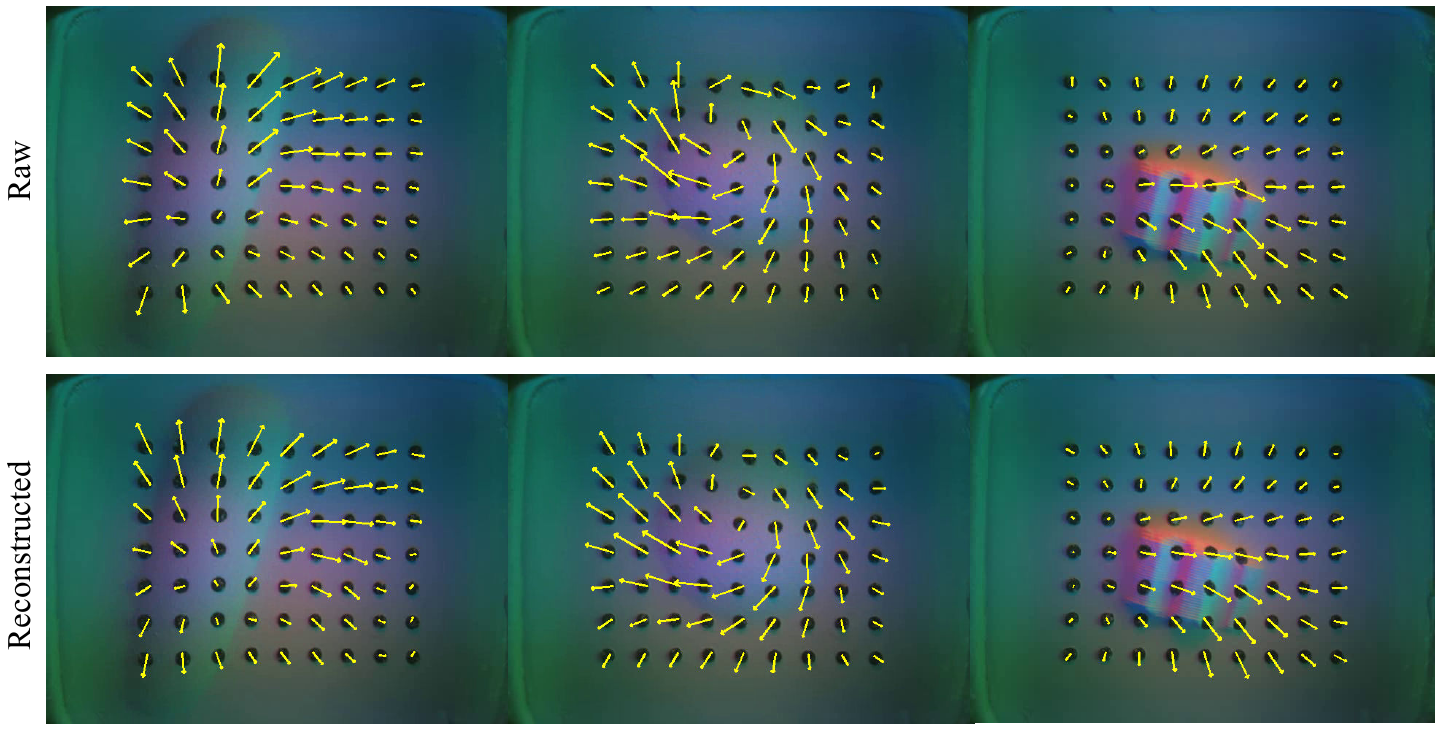}
    \caption{Examples of PCA reconstruction results for marker deformation field of GelSight Mini\cite{gelsight_mini}. The raw tactile images are showed as the background.}
    \label{fig:pca_reconstruction}
\end{figure}

\begin{figure}[!ht]
    \centering
    \includegraphics[width=\linewidth]{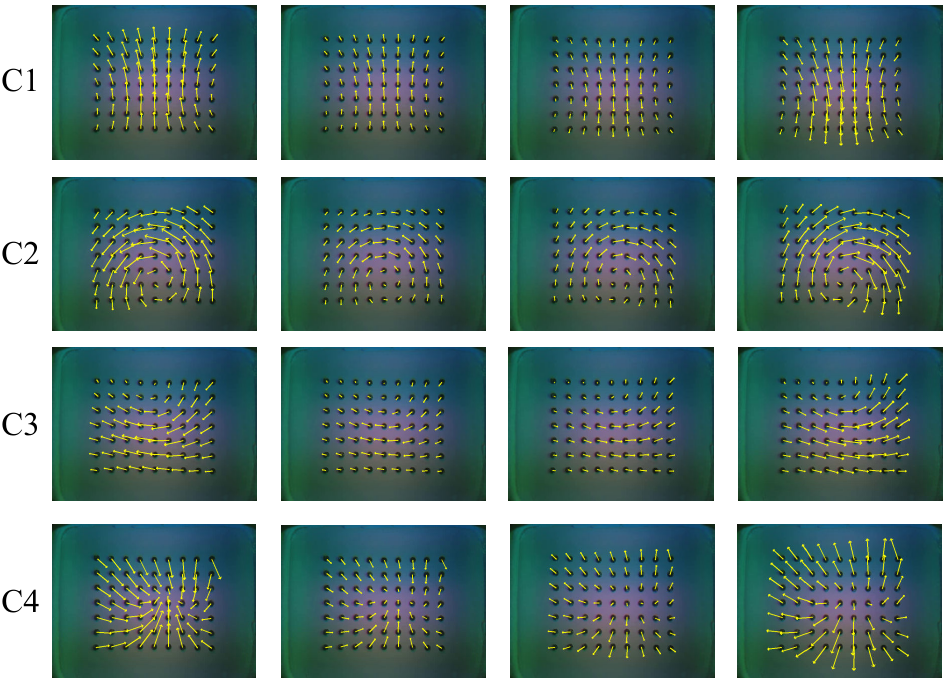}
    \caption{The first four main components (C1-C4) of PCA in our tactile representation. These components are corresponding to tangential force (C1, C3), torsional torque (C2) and normal force (C4) respectively.}
    \label{fig:pca_explain}
\end{figure}

According to \citet{zhang2019effective}, the deformation field of optical tactile sensors can be decomposed into several independent components that are highly explicable (\eg normal force, torsional force \etc). Inspired by this, we use Principal Component Analysis~(PCA) to reduce the dimension for the deformation field $F$. Specifically, we collect a small tactile dataset $\mathcal{D}_{tactile}$ of random interaction with different objects and calculate the deformation field matrix $F_t$ for each frame  $t$ in $\mathcal{D}_{tactile}$. 
Please see the Appendix \ref{sec:supp_dataset} for more details of $\mathcal{D}_{tactile}$. The dataset $\mathcal{D}_{tactile}$ has $m$ frames, and we concatenate and reshape the deformation field $F_t$ in all frames as a new matrix $F_{concat}$ with shape $(m, 2n)$.  We perform PCA on $F_{concat}$ and get the matrix after dimensionality reduction $F_{concat}^{reduced}=T_{proj}  F_{concat}$ with shape $(m, d)$, where $T_{proj}$ is the PCA projection matrix and $d$ is the number of main components. When encountering a new frame $F_{t'}$, we can get the $d$-dim tactile PCA feature by $f_{t'}^{reduced}=T_{proj}F_{t'}$. In practice, we find $d=15$ is enough for reconstruction without losing much details (see Fig. \ref{fig:pca_reconstruction}). We also visualize the first few main components of PCA in Fig. \ref{fig:pca_explain} and surprisingly find that these components are highly explicable and represent the force/torque of different directions.  The use of PCA feature makes the model more robust to tracking errors and noise of marker deformation field. The interpretability of PCA feature also allows the policy network to utilize the tactile representation more easily.

\subsection{Details of the Data Collection Process}\label{sec:supp_dataset}
\begin{figure}[!ht]
    \centering
    \includegraphics[width=\linewidth]{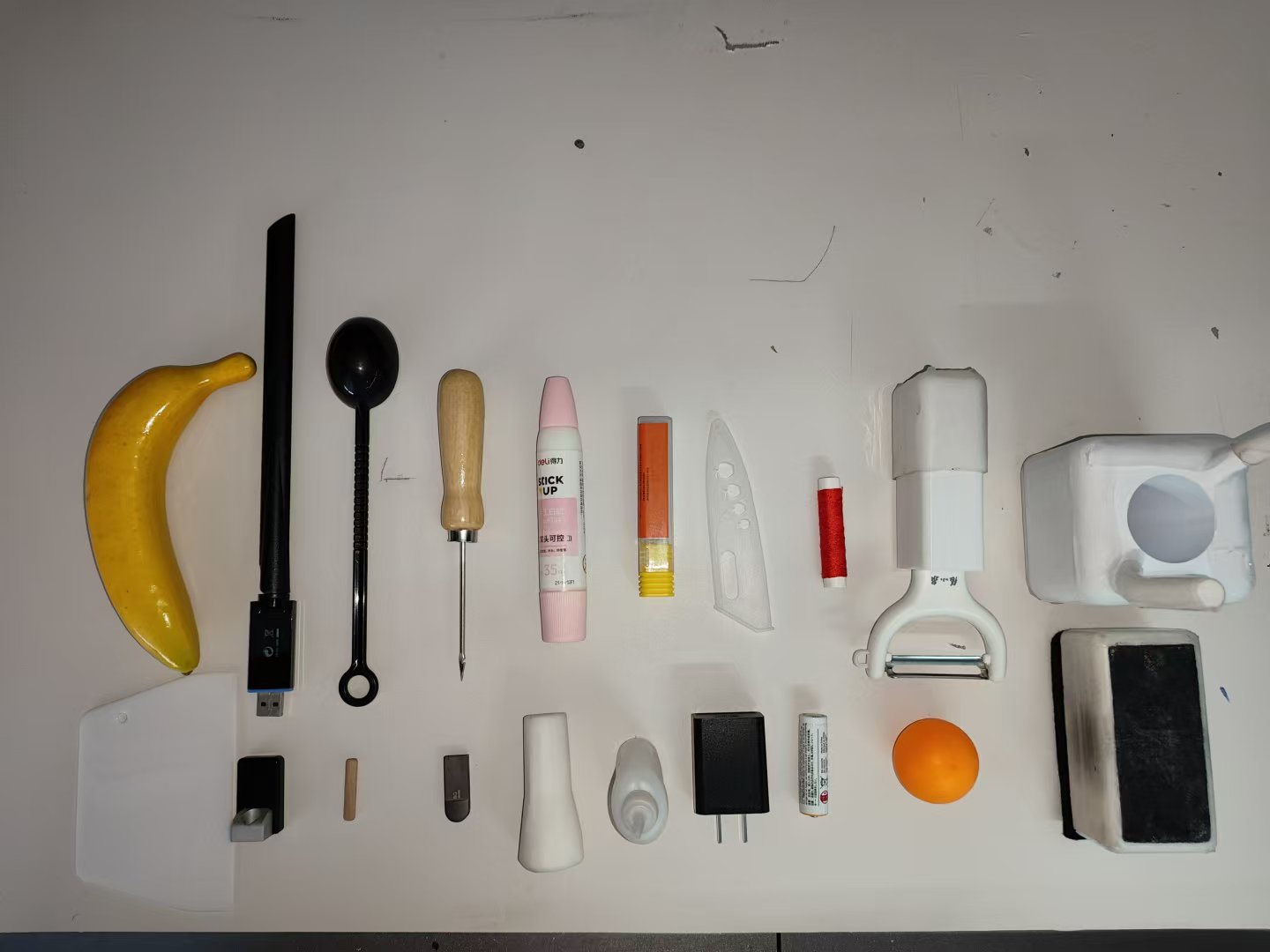}
    \caption{The objects used in the tactile dataset for PCA embedding.}
    \label{fig:tactile_dataset}
\end{figure}

For the tactile dataset used for PCA embedding,
we collect 30 minutes of random interaction data with Gelsight Mini \cite{gelsight_mini} and 40 minutes of random interaction data with MCTac \cite{mctac} with 20 objects in Fig. \ref{fig:tactile_dataset}. 

For three tasks used in our experiments, we collect 60 demonstrations for \textit{Peeling} task, 80 demonstrations for \textit{Wiping} task and 50 demonstrations for \textit{Bimanual Lifting} tasks with \teleop system. During the data collection process, we proactively recorded some reactive behaviors to enhance the robustness of the model. 
\subsection{Details of the Evaluation Protocols}\label{sec:supp_evaluation}
Due to the involvement of human interaction during testing, we employed a \textbf{single-blind} testing methodology to minimize the impact of subjective judgment. In each experiment, a model is randomly selected for evaluation, and the evaluator is unaware of which model is being assessed each time.

\subsection{Details of the Inference Process}\label{sec:supp_impl_details}
We use observation $T_o=2$ for all Diffusion Policy baselines and our Latent Diffusion Policy (LDP). The Diffusion Policy baselines predict open-loop 12FPS action sequences. The default action chunk size is about 0.67s in real-world time. The slow policy will periodically (1-2Hz) predict new inference results at time intervals determined by the action chunk.  The fast policy (AT) takes tactile / force observations at 24 FPS and output action predictions at 24 FPS. The actions are interpolated and sent to robots with a higher frequency ($>$500Hz) with Flexiv RDK \cite{flexiv_rdk}.

\subsection{More Ablation Results}\label{sec:supp_more_ablation_results}
To validate the necessity of shear and torsional forces, we conduct a ablation experiment by removing the shear force in the data and only keeping the normal force dimension (the same as 3D-ViTac \cite{huang20243dvitac}). As shown in Table~\ref{tab:rebuttal_results}, using only the normal force significantly decreases the policy performance ($0.95 \rightarrow 0.48$).

To test the contribution of the asymmetric tokenizer (AT), we design 
a ablation experiment by concatenating action and tactile / force inputs together into the encoder. As shown in Table~\ref{tab:rebuttal_results}, the average score of the policy decreases (0.95 $\rightarrow$ 0.58), demonstrating the effectiveness of AT.

\begin{table}[!h]
\vspace{-2mm}
    \centering
\caption{RDP (Force) Performance for Peeling Task}
\label{tab:rebuttal_results}
\resizebox{\linewidth}{!}{
    \begin{tabular}{l|ccc|c}
        \toprule
         & No& Perturb.&Perturb.& All \\
 & Perturb.& before Contact& after Contact&\\
        \midrule
        \textcolor{gray}{DP} & \textcolor{gray}{0.56} & \textcolor{gray}{0.58} & \textcolor{gray}{0.19} & \textcolor{gray}{0.44}\\
        \midrule
        RDP w. only normal force & 0.47 & 0.62 & 0.36 & 0.48\\
        RDP w. symmetric tokenizer & 0.75 & 0.62 & 0.36 & 0.58\\
        \midrule
        RDP (default) & \textbf{0.99} & \textbf{0.98} & \textbf{0.88} & \textbf{0.95} \\
 \bottomrule
    \end{tabular}
    \vspace{-8mm}
}
\end{table}

\subsection{Details of the TactAR User Study}\label{sec:supp_user_study_details}
 We invited 10 users with different levels of experience in VR teleoperation or Imitation Learning (IL). Each pair of users participated in experiments on the  \textit{Peeling} task, whereby one user teleoperated the robotic arm to peel a cucumber, while the other held the cucumber. The teleoperator was subjected to two distinct settings: traditional VR teleoperation and TactAR. For each setting, they conducted 10 peeling trials, with the cucumber pose varying in each trial. We recorded the length of the cucumber peel obtained in each trial and assessed whether the force applied by the robotic arm was consistently stable from the perspective of the user holding the cucumber. We perform $10\times10\times2=200$ trials for quantitative analysis.

\subsection{Details of the Hyperparameters}\label{sec:supp_hyperparam}
We list the main hyperparameters of RDP and DP in Tab. \ref{tab:hyperparams}. Please see our open-sourced code for other hyperparameters. 
 \begin{table}[!ht]
    \centering
\caption{Hyperparameters for RDP and DP.}
\label{tab:hyperparams}
\begin{tabular}{l|l|c}
    \toprule
    Model & Parameter & Value \\
    \midrule
    \multirow{8}{*}{AT}
    & prediction horizon& 32 @ 24Hz \\ 
    & downsample ratio & 4 \\
    & hidden dim. & 32 \\
 & latent action dim.&4 (peeling), 8 (wiping \& lifting)\\
    & training epoch & 600 \\ 
    & max learning rate & $10^{-3}$ \\
    & weight decay & $10^{-4}$ \\
    & KL penalty multiplier & $10^{-6}$ \\
    \midrule
    \multirow{9}{*}{LDP}
    & observation horizon & 2 @ 12Hz \\
 & proprioception type&relative (peeling \& wiping), \\
 &  & absolute (lifting) \\
    & image resolution & 240$\times$320\\
    & kernel size of 1D U-Net & 3 \\
    & hidden dim. of 1D U-Net & 512, 1024, 2048 \\
    & training epoch & 400 \\
    & max learning rate & $10^{-4}$ \\
    & weight decay & $10^{-6}$ \\
    & training diffusion iter. & 100 \\
    & inference diffusion iter. & 8 \\
    \midrule
    \multirow{3}{*}{DP}
    & prediction horizon& 16 @ 12Hz \\
 & proprioception type&absolute\\
    & kernel size of 1D U-Net & 5 \\
    & training epoch & 600 \\
    & others & the same as LDP \\
 \bottomrule
\end{tabular}
\end{table}

\end{document}